\documentclass[%
 reprint,
 aps,
 pre,
 amsmath,amssymb,
 floatfix,
 superscriptaddress
]{revtex4-2}

\usepackage[english]{babel}

\usepackage{graphicx}
\usepackage{xcolor}
\usepackage[mathscr]{euscript}
\usepackage[colorlinks=true,allcolors=blue]{hyperref}

\begin{document}

\title{Scalar Representations of Neural Network Training Dynamics}

\author{Pedro Jiménez-González}
\author{Miguel C. Soriano}
\author{Lucas Lacasa}
\affiliation{Institute for Cross-Disciplinary Physics and Complex Systems (IFISC, CSIC-UIB), Campus UIB, 07122 Palma de Mallorca, Spain}

\begin{abstract}
Training in artificial neural networks can be viewed as a trajectory evolving through a high-dimensional loss landscape. However, the large number of trainable parameters makes the direct analysis of these dynamics challenging. In this work, we treat such training trajectories as temporal networks and apply recently proposed strategies for the scalar embedding of temporal networks. We investigate whether such a scalar embedding provides a meaningful low-dimensional representation of neural network training dynamics. Using a multilayer perceptron trained on the MNIST classification task, we show that the embedding preserves the main dynamical features observed in the original parameter space, including the emergence of sensitivity to initial conditions for specific learning rate regimes and an accurate reconstruction of the network's maximum Lyapunov exponent. We then use the embedded scalar trajectory to define a characteristic time, analogous to a Lyapunov time, after which the exponential separation between initially close embedded trajectories saturates. This characteristic time captures the typical decorrelation time between initially close network trajectories in the original high-dimensional system. Finally, we investigate the statistical organization of asymptotic training states through a spacing observable defined in the embedded space. We find that the distributions of rescaled asymptotic spacings collapse onto a common form across initial conditions and are compatible with a skew lognormal distribution. Altogether, our results suggest that scalar low-dimensional embeddings provide a useful framework for studying and visualizing the dynamical properties of neural network optimization trajectories. 
\end{abstract}

\maketitle

\section{Introduction}

%Intro general de neural networks desde la optica de la fisica y los sistemas dinamicos
Artificial neural networks (ANNs) are foundational architectures in machine learning and deep learning \cite{goodfellow2016deep}. While possibly originating and being developed within computer science, very early on \cite{hopfield1982neural} and throughout the last decades ANNs have been studied through the lenses of physics \cite{carleo2019machine} and dynamical systems \cite{brunton2022data}, giving rise to three main lines of research. First, the loss functions used to train ANNs often admit a Hamiltonian interpretation, allowing ANN optimization to be framed as the search for a system’s ground state; this connection has driven extensive work linking neural networks to the statistical physics of disordered systems \cite{pei2026statistical}. Second, ANNs are inherently dynamical, exhibiting complex behavior both during training and at inference time, which motivates analyses grounded in dynamical systems theory \cite{brunton2022data}. Third, the evolving structure of a network --from initialization through training to its final optimized form-- encodes information about its function; this ``structure versus function'' viewpoint has been fruitfully explored using tools from network science \cite{kaviani2021application}.

%Nos fijamos en el entrenamiento. Concepto de network trajectories y su relacion con temporal networks
\medskip \noindent
An archetypal example of an ANN is the multilayer perceptron (MLP) \cite{goodfellow2016deep} or feed-forward network, consisting of stacked layers of artificial neurons densely linked between them that represent overparametrised nonlinear input-output functions. The training of an MLP (i.e. the iterative search for the best values of the parameters of the input-output function that fulfils a given task) can itself be graphically represented as a time series of different graph structures, where each graph snapshot at time $t$ represents the updated structure of the MLP (including its weights and biases) at that particular time step of the optimization process. In other words, the optimization process of an MLP can be seen as a trajectory in graph space. Interestingly, in the last years the concept of {\it network trajectories} \cite{lacasa2022correlations} has been coined in network science \cite{latora2017complex}, as an interpretation of the more general concept of {\it temporal networks} \cite{masuda2016guide, holme2019temporal} that emerges from a dynamical systems viewpoint when a temporal network is interpreted as sampling a latent graph dynamics. Recent works describing network trajectories include characterization of network correlations \cite{lacasa2022correlations, caligiuri2025characterizing, bauza2023characterization, hartle2025autocorrelation} or sensitivity to initial conditions \cite{caligiuri2023lyapunov, caligiuri2026predictability}.

%Resultados interesantes desde esta optica, trayectorias caoticas y demas
\medskip \noindent 
This perspective of network trajectories has been used to assess the training of neural networks, viewing training as tracing trajectories in graph space \cite{danovski2024dynamical}, rather than the more traditional approach of focusing solely on the evolution of the loss function and the geometry of the loss surface. 
This approach has unveiled the onset of complex dynamics emerging in gradient-descent optimization, including seemingly universal (i.e. architecture-independent) transition to chaotic behavior at large learning rates of the gradient descent optimizer, along with evidence of training optimality close to the onset of chaos \cite{jimenez2026leveraging}, in line with broader observations that neural systems may benefit from operating near critical or edge-of-chaos regimes \cite{morales2021optimal}.
Other dynamical-systems approaches have characterised network's spatial expressivity and decision boundaries in the input data \cite{storm2024finite}, used Lyapunov-based indicators to quantify stability in recurrent neural architectures \cite{gallicchio2018local}, or studied how the geometry of the loss landscape and the distribution of minima influence optimization dynamics and learning outcomes \cite{bosman2020visualising}. 
These studies reinforce the view that neural network training should be understood as a dynamical process unfolding in a high-dimensional landscape.\\
%Resultados clasicos de analizar la optimizacion, y los embeddings, que eventualmente nos dejan visualizar. Interpretando la evolucion de la red neuronal como una red temporal, abre la puerta a usar resultados de embedding de redes temporales.
Observe, however, that characterising network trajectories of realistic MLPs is challenging due to their extremely high dimensionality: even a toy MLP can already contain thousands of trainable parameters (links). As training progresses, these parameters collectively trace a trajectory through a highly non-convex loss landscape, where each point represents a particular configuration of the network. Although the optimization algorithms used are designed to minimize loss, the actual path taken through this landscape reflects a complex interplay between gradient directions, learning rate, model architecture, and data structure. Capturing and {\it visualizing} this high-dimensional trajectory is therefore both a technical and conceptual challenge, and a better understanding of these high-dimensional network trajectories can offer valuable insights into the explainability of ANNs.
Early works aiming to understand optimization in deep networks focused on probing the geometry of such high-dimensional loss landscape through low-dimensional slices and interpolation experiments. For instance, the linear path experiment introduced in \cite{goodfellow2014qualitatively} analyzes the loss along a straight segment between initialization and a trained solution, and helped popularize the idea that large-scale barriers may be limited in certain settings. Subsequent work developed more systematic visualization techniques and reported links between landscape geometry and generalization, including methods to obtain comparable 2D loss plots \cite{li2018visualizing} and analyses connecting sharpness to generalization performance \cite{wu2017towards, ghorbani2019investigation, sagun2017empirical}. At the same time, recent studies have cautioned that conclusions drawn from 1D/2D slices can be fragile: the behavior of linear interpolation depends on the task, architecture, initialization, and training setup, and modern networks can exhibit qualitatively different interpolation profiles \cite{frankle2020revisiting, vlaar2022can, lucas2021analyzing}. More recently, theoretical frameworks have proposed that deep-learning loss landscapes may exhibit a multifractal structure \cite{ly2025optimization}.\\
Another theme emerging from these lines of research is that training dynamics and solution sets can display an effective low-dimensional structure, and thus dimensionality reduction of the blackbox training process is sensible. This perspective is supported by work on intrinsic dimension and subspace training \cite{li2018measuring}, by observations that gradients align with structured low-dimensional subspaces during training \cite{gur2018gradient, song2024does}, and by evidence that trajectories of trained models concentrate on low-dimensional manifolds in prediction space across architectures and training choices \cite{mao2024training}. Altogether, these results motivate studying training --i.e. network trajectories-- through compact representations that preserve relevant geometric and dynamical information. For instance, some authors have proposed to directly project and visualize network trajectories by applying dimensionality reduction to snapshots of weights collected during training: \cite{lorch2016visualizing} projects training trajectories using PCA, and \cite{antognini2018pca} compares the resulting projections with those of high-dimensional random walks, highlighting characteristic low-dimensional patterns. More broadly, recent work treats neural weights as data objects and learns representations in weight space for analyzing and comparing models \cite{eilertsen2020classifying, schurholt2021self, han2026survey}. These approaches suggest that embedding network trajectories can offer a complementary lens on optimization and generalization. Interestingly, from the field of temporal networks some authors have indeed proposed recently a range of embedding techniques, to effectively project the trajectory of generic evolving networks into trajectories in lower dimensional space \cite{thongprayoon2023embedding, lacasa2025scalar, lacasa2026fluid}.

%Nuestra propuesta: vamos a analizar un embedding concreto (escalar), y en particular nos interesa ver si ese embedding es capaz de preservar dinámicas no triviales. Para ellos consideramos el caso del PRR, en donde observamos una transición al caos en las network trajectories. Decimos lo que hacemos.
\medskip
In this work we contribute to the task of building low-dimensional embeddding of (neural) network training trajectories. To that aim, we consider the scalar embedding technique proposed in \cite{lacasa2025scalar}, and assess to what extent the complexity of high-dimensional network trajectories is preserved in the scalar projection. In particular, we consider the transition-to-chaos phenomenology evidenced by high-dimensional deep neural network trajectories described in \cite{jimenez2026leveraging}. We show that the properties of the high-dimensional dynamics --including the transition to chaos, and the onset of positive network Maximum Lyapunov Exponent nMLE--, are preserved and well captured already in the scalar projection of the network trajectories. After such validation, we explore the properties of these scalarized network trajectories, finding relevant insights of the original high-dimensional dynamics, including a simple characterization of the network decorrelation times and a nontrivial spacing distribution of  local minima of the loss landscape.\\
The rest of the paper is as follows: in section \ref{sec:scalar-embedding} we make a brief summary of the scalar embedding methodology. Section \ref{sec:results} depicts the results of this work. After providing details on the system under analysis, we present results pointing to the emergence of a transition-to-chaos phenomenology in the original high-dimensional neural network trajectory, and find that the scalar embedding of such trajectory fully preserves such phenomenology. Subsequently, we show that the decorrelation time --i.e. the characteristic time after which two initially close neural networks with chaotic evolution decorrelate under the action of the gradient descent dynamics-- can be extracted from the scalar embedding. We finally explore the distribution of local minima in the scalar embedding, finding evidence of a universal spacing distribution. In section \ref{sec:conclusion} we conclude. 

%adopt such a perspective by embedding the evolution of network parameters into a low-dimenisonal representation, enabling us to study the organization of minima and the statistical structure of the landscape explored during training. 
\begin{figure*}[htb!]
    \centering
    \includegraphics[width=0.8\textwidth]{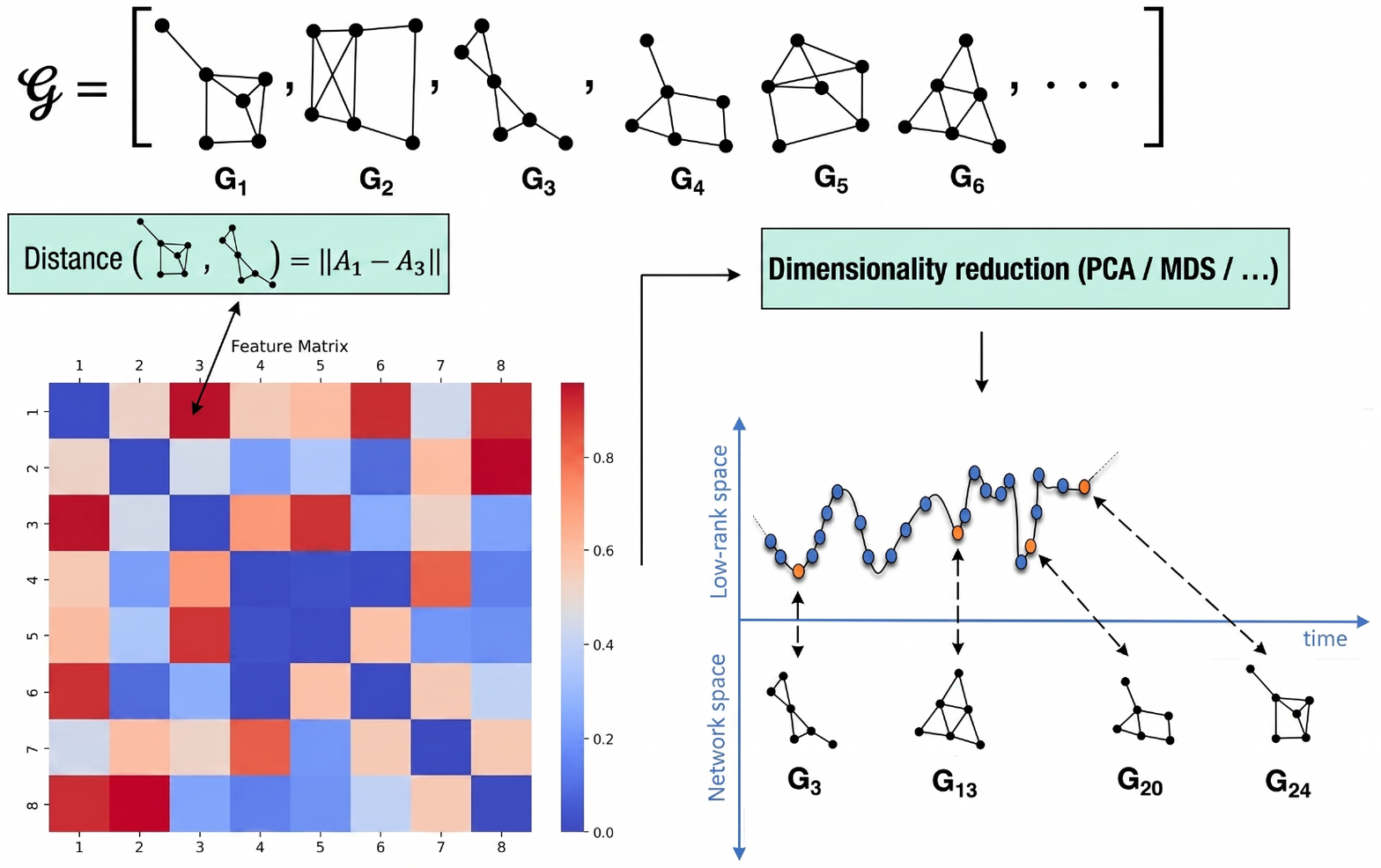}
    \caption{Sketch of the methodology to extract a scalar embedding of a temporal network, based on the methodology in \cite{lacasa2025scalar}. A temporal network trajectory is given by an ordered sequence of graph snapshots. Each graph snapshot is then mapped into feature space, where the features of a given graph snapshot are the relative distances between such snapshot and any other snapshot. The resulting feature matrix is finally dimensionally reduced to obtain a low-dimensional representation of the original trajectory.}
    \label{fig:cartoon}
\end{figure*}

\section{Scalar embedding of network trajectories}\label{sec:scalar-embedding}
In this work, we employ the scalar embedding methodology in \cite{lacasa2025scalar}, which characterizes the dynamics of temporal networks through low-dimensional representations. For illustration, a sketch of the methodology is depicted in Fig.~\ref{fig:cartoon}.

\medskip
\noindent A temporal network, or network trajectory, is defined as an ordered sequence of graphs $\mathcal{G}=(G_{1}, G_{2},...,G_{T})$, where each $G_{t}$ is the network snapshot at time $t$, represented by its adjacency matrix $A(t)$.
The key idea is that, in order to preserve the key dynamical properties of a temporal network, rather than any other microscopic property of each network snapshot, one should aim to preserve the relative position between network snapshots in a low-dimensional representation $\mathcal{G}_{\phi}=(z_{1}, z_{2},...,z_{T})$, where $z_{t} \in \mathbb{R}^{\text{dim}}$, with $\text{dim} \ll T$. Accordingly, each network snapshot in the temporal network is initially mapped into a point in a feature space, where the features account for the relative distance between such snapshot and every other snapshot in the temporal network, and discards other features with e.g. topological information of the snapshots. 
The feature embedding is constructed by first defining pairwise distances between snapshots. Specifically, each snapshot $G_{t}$ is represented by a feature vector composed of its squared distances to all other snapshots along the trajectory. For any two graph snapshots $G_{t}, G_{t'}$, its distance is computed as $d(G_t, G_{t'}) = ||A - A^{'}||_{2}$, where $A$ and $A'$ are the adjacency matrices of  $G_{t}$ and $G_{t'}$, respectively. Collecting these square distances yields a feature matrix $\mathcal{D}^{(2)}=\{d^2_{tt'}\}_{t,t'=1}^T$. Observe that the number of features of each network snapshot only depends on the number of time snapshots, rather than the size of each network snapshot (i.e. the number of nodes).
Subsequently, one proceeds by dimensionally reducing the feature space by maximally preserving the relative distance between points. A variety of different possibilities can be conducted \cite{lacasa2025scalar}, here we make use of classical multidimensional scaling \cite{mead1992review}. 
Finally, a low-rank representation of the temporal network is found. Of particular interest is the case $\text{dim}=1$, which produces a scalar embedding that compresses the temporal evolution of the entire network into a real-valued univariate time series.

\medskip \noindent 
In \cite{lacasa2025scalar} this methodology was validated across a range of synthetic and empirical network trajectories with different dynamics, demonstrating that the resulting scalar embeddings preserve key dynamical properties of the original trajectories across different types of dynamical complexity, and for a large range of network sizes. In the following, we adopt this framework to analyze the training dynamics of artificial neural networks. Notice that it is not obvious a priori if such approach can preserve dynamics, as in deep learning even small neural networks are really large in terms of the number of edges (training parameters).

%For further methodological details, we refer the reader to the original work. 

\section{Results}\label{sec:results}
To explore the ability of the embedding methodology to capture relevant neural network dynamics, we initially focus on the main findings reported in \cite{jimenez2026leveraging}. In that work we showed that, for a range of learning rate values, the training trajectories of neural networks exhibit a transition to chaos, as characterized by positive Maximum Lyapunov Exponents (MLEs). 
Although \cite{lacasa2025scalar} demonstrated the validity of the approach on both synthetic and empirical networked systems, those systems are significantly smaller and structurally simpler than modern neural architectures. Therefore, as a first step, it is essential to assess whether applying a scalar embedding to a neural network trajectory can still preserve meaningful dynamical information, or whether such an extreme dimensionality reduction might lead to the loss of relevant structure, thereby limiting the interpretability of the embedding.

\subsection{Background: transition to chaos in multilayer perceptron training dynamics}\label{sec:background}
Our validation builds upon the theoretical framework introduced in \cite{jimenez2026leveraging}, which analyzes neural network training from a dynamical-systems perspective through the evolution of their weight trajectories. In that work, we quantified how small perturbations in the initial conditions lead to diverging network evolutions, establishing a connection between learning rate, training stability, and chaotic dynamics. Motivated by these findings, we interpret the evolution of the network parameters as a temporal dynamical system, which can be embedded into a low-dimensional representation. This approach provides the basis for assessing whether the scalar embedding captures the same dynamical features ---most notably, sensitivity to initial conditions--- identified in the original high-dimensional system.\\
We briefly summarize the neural network model used in \cite{jimenez2026leveraging}, which serves as the experimental setup for the present study. We consider a multilayer perceptron (MLP), i.e., a fully connected feed-forward network with a single hidden layer, trained on the supervised classification task MNIST \cite{MNISTdataset}. An MLP implements a nonlinear mapping $y=\mathscr{F}(x;\Omega)$, where $x \in \mathbb{R}^m$ is the input, $y$ is the output, and $\Omega = \{\omega_k\}$ denotes the set of trainable parameters (weights and biases). In our example of interest, the architecture of the MLP consists of an input layer with 784 neurons, a hidden layer with 64 neurons, and an output layer with 10 neurons, resulting in approximately $5\cdot 10^{4}$ trainable parameters. Hidden layers perform affine transformations followed by a tanh nonlinear activation function, while the output layer produces the final prediction.\\
Training is formulated as the minimization of a cross-entropy loss function $\mathcal{L}(x;\Omega)$ through full-batch gradient descent (GD), according to
\begin{equation}
    \omega_{k}(t+1) = \omega_{k}(t) - \eta \partial_{\omega_{k}}\mathcal{L}(x;\Omega)
\end{equation}
where $\eta$ denotes the learning rate. The sequence of parameter configurations updates
\begin{equation}
    \mathcal{T} = \{\Omega(0), \Omega(1), \Omega(2), \ldots\},
\end{equation}
with $\Omega(t) = \{\omega_1(t), \ldots, \omega_m(t)\}$, defines a trajectory in parameter space and denotes the set of trainable parameters at training epoch $t$. This trajectory describes how the network evolves and explores the high-dimensional loss landscape during training, and constitutes the object of analysis in this work.

\begin{figure}[h]
    \centering
    \includegraphics[width=1\linewidth]{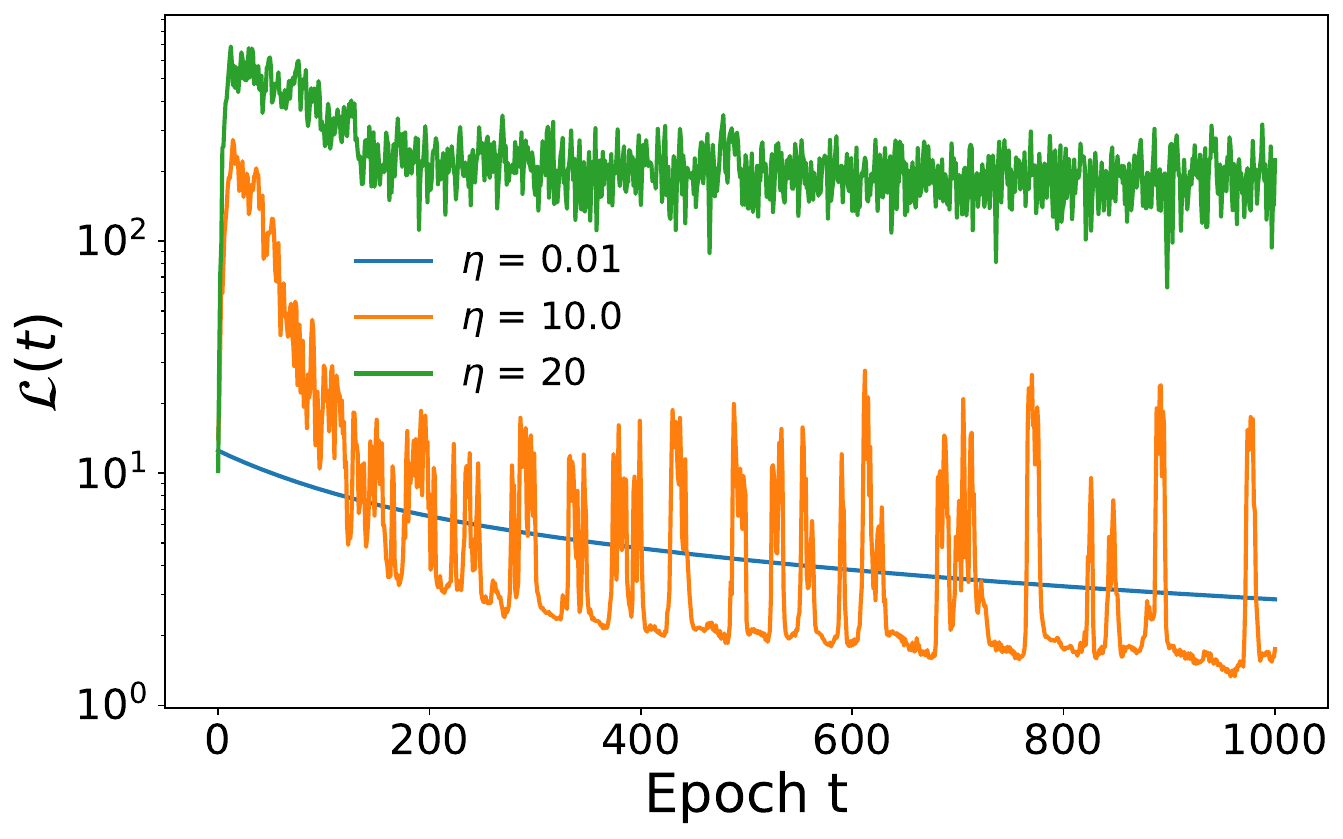}
    \caption{Training loss trajectories for three representative learning rates $\eta$. Full batch training is performed at each epoch.}
    \label{fig:loss}
\end{figure}

\medskip \noindent
The dynamical analysis performed in \cite{jimenez2026leveraging} revealed the existence of three distinct training regimes as a function of the learning rate, illustrated in Fig~\ref{fig:loss}. Small values of $\eta$ produce smooth and monotonic convergence of the loss function, intermediate values generate irregular but still convergent transient dynamics, and sufficiently large learning rates lead to strongly irregular behavior where learning fails.  

To characterize the dynamical properties of the network trajectories along training and their sensitivity to initial conditions, we considered pairs of networks initialized with infinitesimally close conditions. Specifically, given a reference parameter set $\Omega = \{w_{ij}\}$, a perturbed configuration is constructed as $\Omega' = \{w_{ij} + \delta_{ij}\}$, where $\delta_{ij}$ are independent and identically distributed (i.i.d.) random variables drawn from a uniform distribution $\hat{\delta} \sim U(-\epsilon, \epsilon)$, with perturbation radius $\epsilon = 10^{-8}$. 

\begin{figure}[h]
    \centering
    \includegraphics[width=1\linewidth]{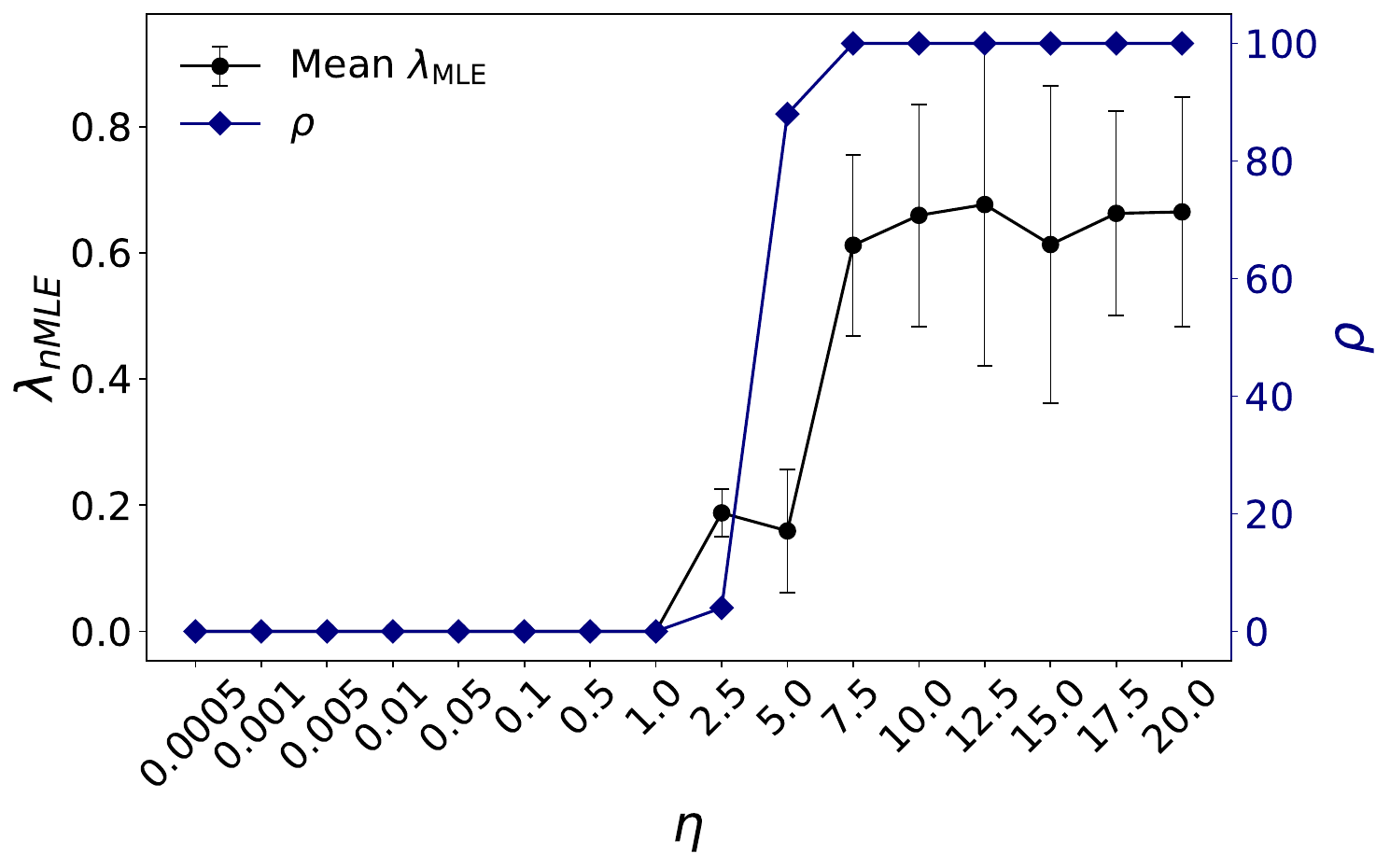}
    \caption{The black curve shows the averaged network Maximum Lyapunov Exponent $(\lambda_{\mathrm{nMLE}}$) as a function of the learning rate $\eta$, with error bars representing the standard deviation of all local finite-time Lyapunov exponents computed across different initializations. The blue curve displays $\rho$, the percentage of positive local Lyapunov exponents across all realizations.}
    \label{fig:lyapunov_otropaper}
\end{figure}

Given the corresponding trajectories $\{\Omega(t)\}$ and $\{\Omega'(t)\}$, their divergence is quantified by a distance
$d(\Omega(t), \Omega'(t)) = \sum_{\omega_k \in \Omega} |\omega_k(t) - \omega'_k(t)|,$
i.e., the $L_1$ norm of the parameter differences. For certain learning rate values, this distance exhibits an initial exponential growth phase, which enables the estimation of finite-time network Lyapunov exponents \cite{caligiuri2023lyapunov}. The network's Maximum Lyapunov Exponent (nMLE) $\lambda_{\text{nMLE}}$ is then obtained by averaging these local exponents over 50 independent initializations, each with five perturbed replicas \cite{caligiuri2023lyapunov, lacasa2025scalar}. 
The resulting dependence of the nMLE on the learning rate is shown in Fig.~\ref{fig:lyapunov_otropaper}. Positive values of $\lambda_{\text{nMLE}}$ indicate sensitivity to initial conditions, thus identifying regimes in which the training dynamics exhibit chaotic behavior. In this figure we also plot, $\rho$, defined as the fraction of initial conditions displaying positive local network Lyapunov exponents. 
Incidentally (see Appendix~\ref{sec:appendix} for details), the value of $\eta$ marking the sharp transition toward $\rho\simeq100\%$ coincides with the one where the average number of epochs required to reach $90\%$ classification accuracy in the test set for the MNIST task is minimized, revealing the existence of a sweet-spot in the training efficient around $\eta \approx 7.5$, where exploitation and exploration strategies are optimally balanced  \cite{jimenez2026leveraging}. 

\subsection{Sensitivity to initial conditions and transition to chaos in the embedded scalar trajectories}
We now investigate whether the scalar embedding strategy depicted in Section~\ref{sec:scalar-embedding} is capable of capturing the dynamical properties previously identified in neural network training. In particular, we focus on the emergence of sensitivity to initial conditions in a MLP trained on the MNIST classification task, using the same architecture described in Section~\ref{sec:background}. 

\begin{figure}[h]
    \centering
    \includegraphics[width=1\linewidth]{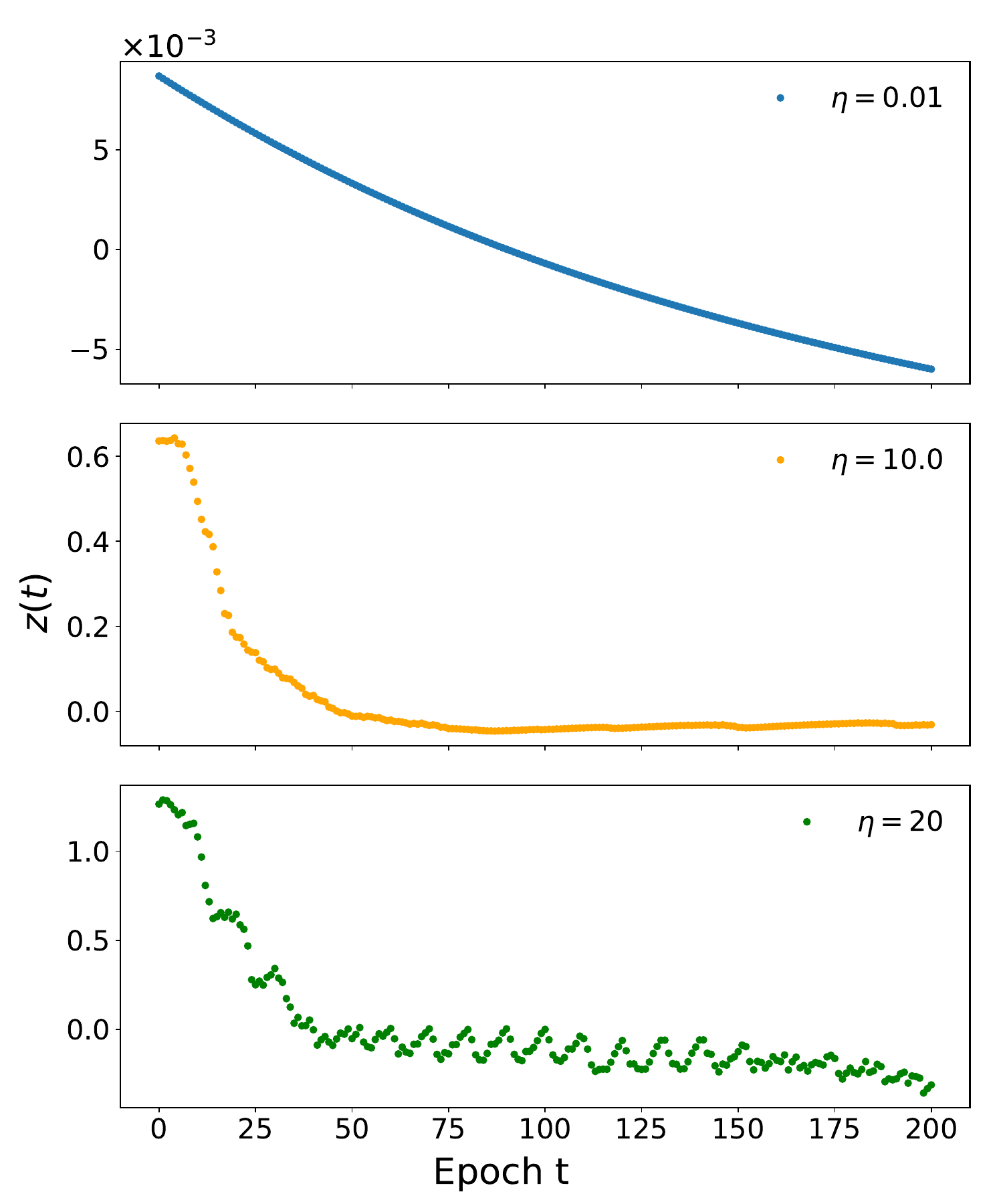}
    \caption{Scalar embeddings of the first 200 training epochs for three different learning rates. Each curve corresponds to the embedded trajectory $z(t)$ associated with a reference network trajectory $\mathcal{G}$.}
    \label{fig:ejemplo-embedding}
\end{figure}

\medskip \noindent 
We generate a reference trajectory ${\mathcal{G} = (\Omega(0), \Omega(1), \ldots, \Omega(T))}$ for the three representative values of the learning rate depicted in Fig.~\ref{fig:loss}. The corresponding embedded trajectories $\mathcal{G}_{\phi}$ over the first 200 epochs are shown in Fig.~\ref{fig:ejemplo-embedding}.
Observe that the scalar embedding preserves the qualitative differences between the training regimes previously identified through the loss evolution in Fig.~\ref{fig:loss}. For small learning rates, the embedded trajectory evolves smoothly and monotonically, reflecting the stable optimization dynamics associated with small learning rates. For intermediate learning rates, corresponding to the regime where sensitivity to initial conditions begins to emerge in the original high-dimensional system, the trajectory remains convergent but evolves on a different scale from the small $\eta$ case, indicating a qualitative change in the underlying training dynamics. Finally, for large learning rates, the embedded trajectory displays fluctuations and irregular behavior. Thus, although the scalar embedding does not provide a one-to-one representation of the loss evolution, nor this is its aim, it successfully captures the fundamental dynamical characteristics of the original training trajectories. This is noticeable given that the embedding projects trajectories evolving in a space of order $5 \cdot 10^{4}$ dimensions onto a single scalar observable, acting as a coarse-grained representation of the underlying parameter space dynamics.

\begin{figure}[h]
    \centering
    \includegraphics[width=1\linewidth]{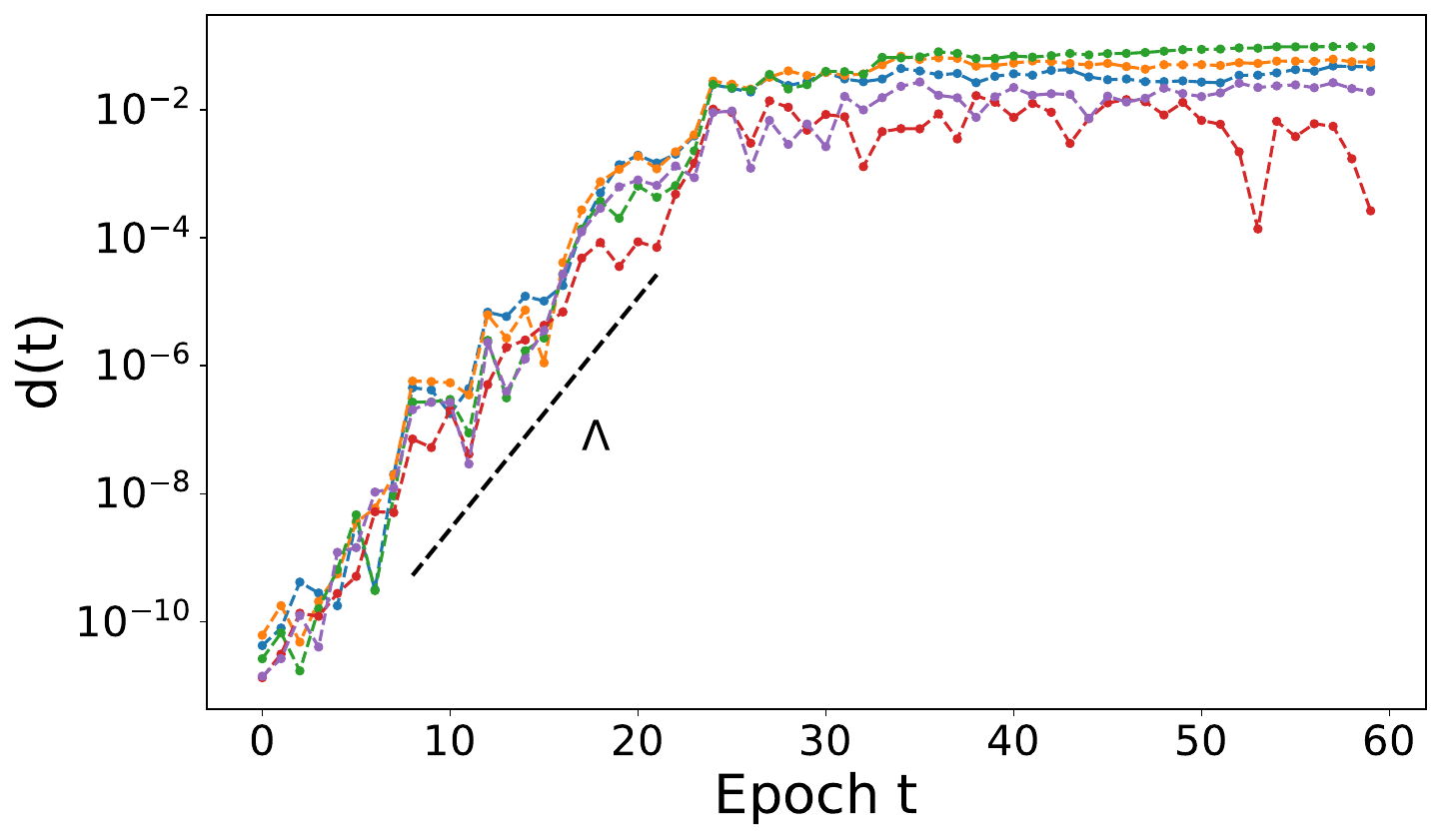}
    \caption{Semi-log plot of the distances $d(t)$ between the scalar embeddings of a reference trajectory and five independently perturbed trajectories, shown as a function of the training epoch $t$ for $\eta = 10$. Each color corresponds to a different perturbation.}
    \label{fig:d(t)-embedding}
\end{figure}

\begin{figure}[h]
    \centering
    \includegraphics[width=1\linewidth]{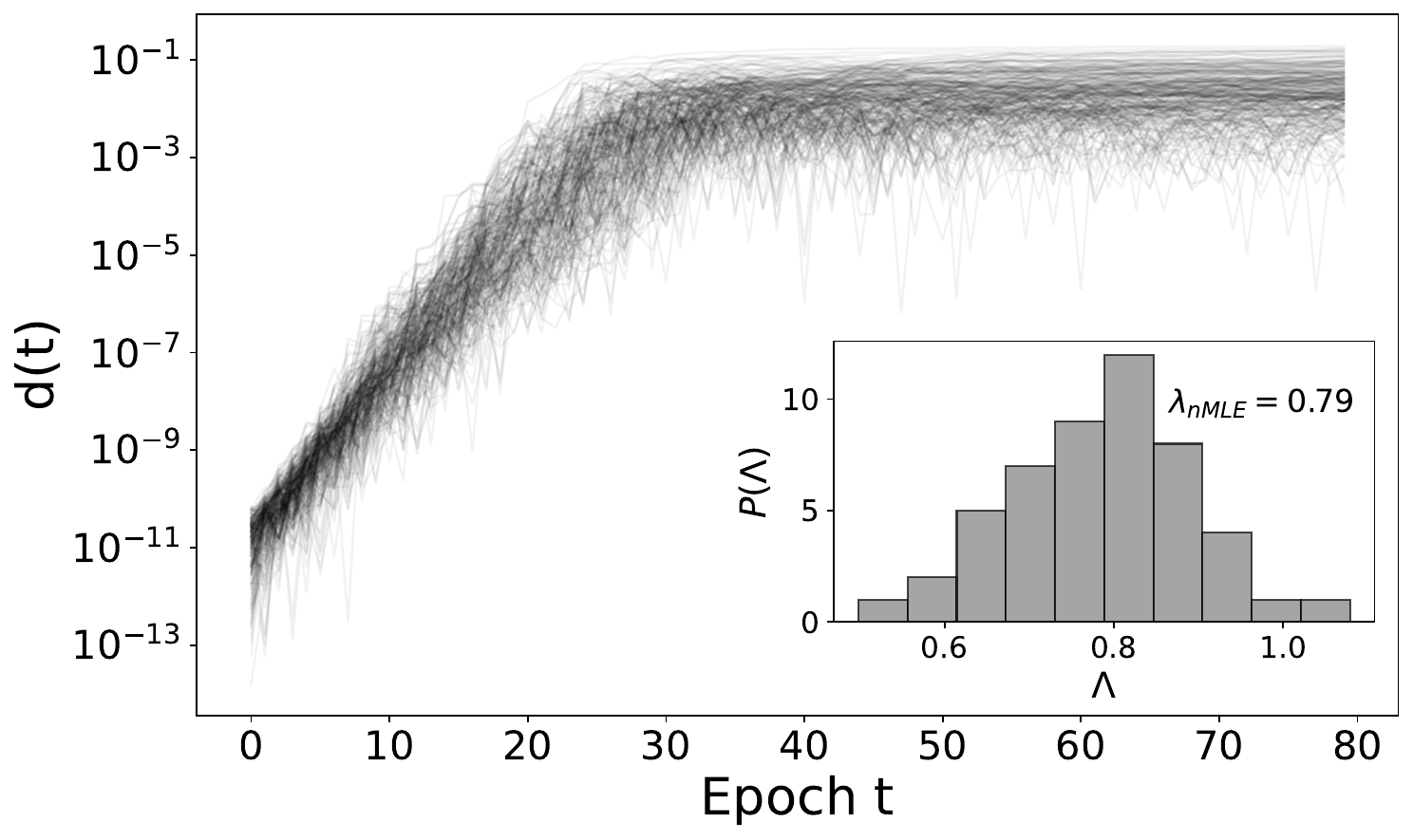}
    \caption{Same as Fig.~\ref{fig:d(t)-embedding} for multiple $\epsilon$-balls centered at different initial conditions $\Omega(0)$. For each initial condition, a local Lyapunov exponent $\Lambda(\Omega(0))$ is extracted from the exponential regime. The inset displays the distribution of these local exponents, whose mean defines the network maximum Lyapunov exponent, $\lambda_{\mathrm{nMLE}}$. }
    \label{fig:fig3}
\end{figure}

\begin{figure}[h]
    \centering
    \includegraphics[width=1\linewidth]{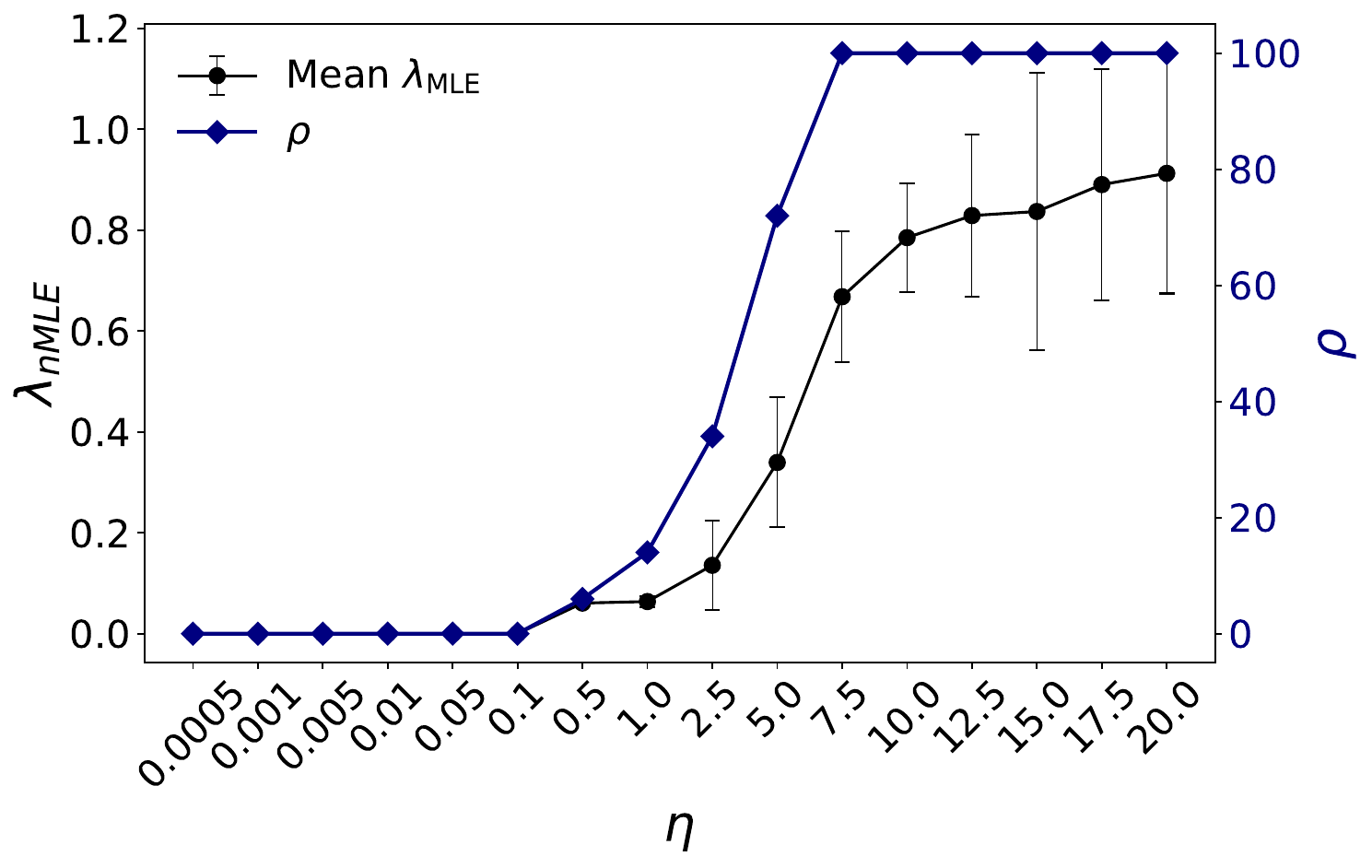}
    \caption{Maximum Lyapunov Exponent estimated from the scalar embedding. The black curve shows the averaged network Maximum Lyapunov Exponent ($\lambda_{\mathrm{nMLE}}$) as a function of the learning rate $\eta$, with error bars representing the standard deviation of all local Lyapunov exponents computed across different initial conditions and perturbations. The blue curve displays $\rho$, the percentage of positive local Lyapunov exponents across all realizations.}
    \label{fig:lyapunov-embedding}
\end{figure}

\medskip \noindent 
We next examine whether the embedding preserves the sensitivity to initial conditions observed in the full parameter space dynamics. Following the procedure described above, for an ensemble of different initial conditions, we generate five perturbations of each initial condition, and compute the $L_{1}$ distance $d(t)$ between the scalar embeddings of the perturbed and reference trajectories. Figure~\ref{fig:d(t)-embedding} shows the resulting evolution for a representative case with learning rate $\eta = 10$, a regime for which sensitivity to initial conditions emerged in the original high-dimensional system. A clear exponential growth phase is observed during the first $\sim 30$ epochs, whose slope defines a local Lyapunov exponent in the embedded space. Then, we repeat the same procedure over an ensemble of 50 independent initial conditions. The resulting ensemble of trajectories is shown in Fig.~\ref{fig:fig3}, where each initial condition gives rise to a potentially different local Lyapunov exponent $\Lambda(\Omega(0))$. The inset displays the distribution of these exponents.\\
Averaging these local exponents yields to the maximum Lyapunov exponent (MLE) $\lambda_{\text{MLE}}$ of the scalar embedded trajectories, which is the scalarized version of the network Maximum Lyapunov exponent $\lambda_{\text{nMLE}}$ previously estimated in the original, high-dimensional system.  At the same time, in the embedded dynamics we can also measure $\rho$, the fraction of initial conditions displaying positive local Lyapunov exponents. Accordingly, these allow for a direct comparison with the result obtained in the full parameter space (Fig.~\ref{fig:lyapunov_otropaper}), now from the perspective of the scalar embedding.
Results are plotted in Fig.~\ref{fig:lyapunov-embedding}, finding a remarkable similarity with the transition to chaos phenomenology depicted in Fig.~\ref{fig:lyapunov_otropaper}. 
%The only noticeable discrepancy is that the embedding method yields positive Lyapunov exponents for $\eta=0.5$ and $\eta=1$, whereas this behavior is absent in the full parameter space analysis. However the corresponding values of $\rho$ indicate that the fraction of positive local exponents in these cases is small, so this discrepancy does not affect the overall agreement between the two approaches. 

%The appearance of positive Lyapunov exponents within the same learning rate regimes demonstrate that the scalar embedding preserves the dynamical signature of sensitivity to initial conditions as in the original high-dimensional space. 
This agreement is particularly remarkable given the drastic dimensionality reduction involved: original trajectories evolving in a space of tens of thousands of parameters are compressed into a single scalar observable while still retaining the instability patterns characteristic of the underlying training dynamics. These results validate the embedding methodology as a reliable tool for capturing non-trivial dynamical properties in artificial neural networks. 

\subsection{Estimating decorrelation times in the original high-dimensional chaotic dynamics from the scalar embedding}

Having shown that the embedding methodology provides a meaningful representation of the training dynamics, we now leverage it to quantify several properties of the chaotic dynamics observed during training. To this end, we now fix the learning rate to $\eta=10$, a representative case within the optimal training efficiency regime, where the dynamics display both strong sensitivity to perturbations and near-optimal training times for the classification task. This choice allows us to characterize the interplay between chaotic dynamics and training efficiency directly in the embedded representation.

\medskip \noindent
In particular we aim to estimate, directly from the embedded trajectories, a quantity that describes the decorrelation time of the original, high-dimensional chaotic dynamics. As a matter of fact chaotic trajectories of initially close conditions fully decorrelate after some time, that we label $\tau_{\mathrm{dec}}$.
%An interesting aspect of interpreting ANN training as a dynamical process is understanding how long two trajectories initialized from nearly identical initial conditions remain effectively close. This decorrelation timescale determines whether the optimization dynamics are robust to small perturbations or instead rapidly evolve toward different regions of parameter space. In a chaotic regime, arbitrarily small perturbations may grow and eventually produce macroscopically distinct trajectories, whereas in a strongly stable regime nearby trajectories may remain close throughout training. 
%Characterizing the timescale associated with this loss of coherence therefore provides a natural way to quantify sensitivity to initial conditions beyond asymptotic indicators such as Lyapunov exponents. 
%To quantify this loss of coherence, we introduce a \emph{decorrelation time}, denoted by $\tau_{\mathrm{dec}}$, defined as the earliest epoch at which two initially nearby trajectories can no longer be regarded as belonging to the same dynamical orbit according to a prescribed criterion. 
In principle, $\tau_{\mathrm{dec}}$ should be computed directly from the evolution of the trajectories in the original parameter space. However, this would require estimating cross-correlation functions between high-dimensional network trajectories evolving in a space of order $5 \cdot 10^{4}$ dimensions, which makes such an approach difficult to implement in practice.\\
Before introducing the protocol to retrieve $\tau_{\mathrm{dec}}$ directly from the scalar embedding, we need a proxy that can be used to validate whether such quantity can indeed be retrieved from the scalar embedding in the first place. Given that it is computationally intensive to estimate cross-correlation functions of high-dimensional dynamics, 
a natural alternative for such proxy is to use the {\it test accuracy} trajectory (i.e. the evolution of the cross-entropy loss computed in the test set), which itself is a scalar magnitude. Note, however, that accuracy constitutes a highly compressed and often saturated projection of the underlying high-dimensional dynamics. Many distinct parameter configurations may yield nearly identical accuracies, and it is often the case that network trajectories which are dynamically varying can yield constant accuracy time series, as happens in the phase before grokking \cite{power2022grokking, clauw2024information}.
%while the corresponding time series typically display plateaus, abrupt transitions, and strong temporal autocorrelations. 
As a consequence, an apparent loss of similarity between two accuracy traces does not necessarily imply that the associated trajectories have evolved toward different regions of the parameter space; it may simply reflect transient differences in optimization speed or alternative pathways within the same basin of attraction. Conversely, trajectories converging to different minima may still exhibit almost indistinguishable accuracy curves if their final performance is comparable. 

\begin{figure}[h]
    \centering
    \includegraphics[width=1\linewidth]{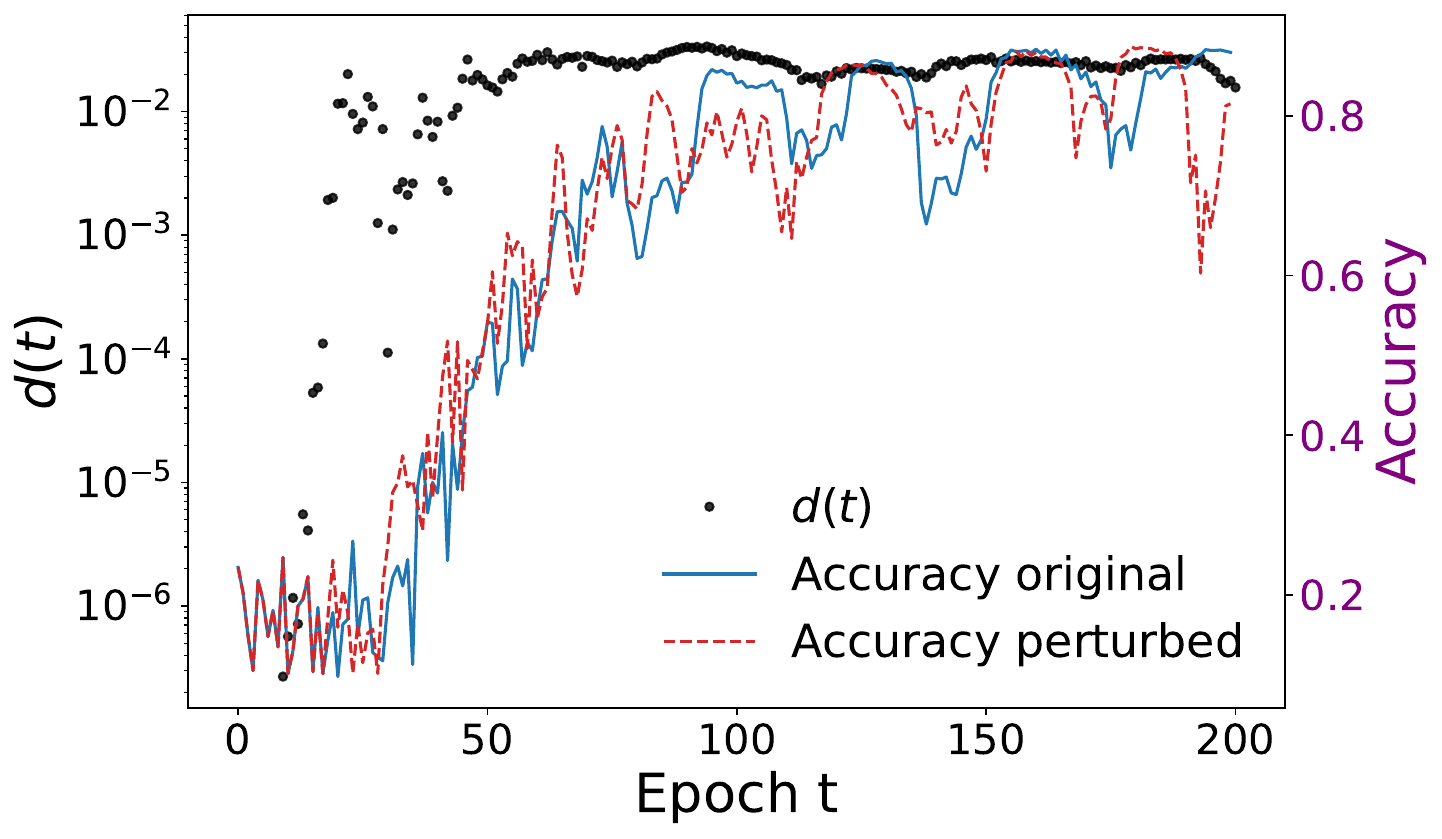}
    \caption{Black markers show the embedding-based separation $d(t)$ between a reference trajectory and one perturbed replica (left axis). The blue solid and red dashed curves show the corresponding accuracy traces (right axis). Results correspond to $\eta=10$.}
    \label{fig:example_accuracy}
\end{figure}

\begin{figure}[h]
    \centering
    \includegraphics[width=1\linewidth]{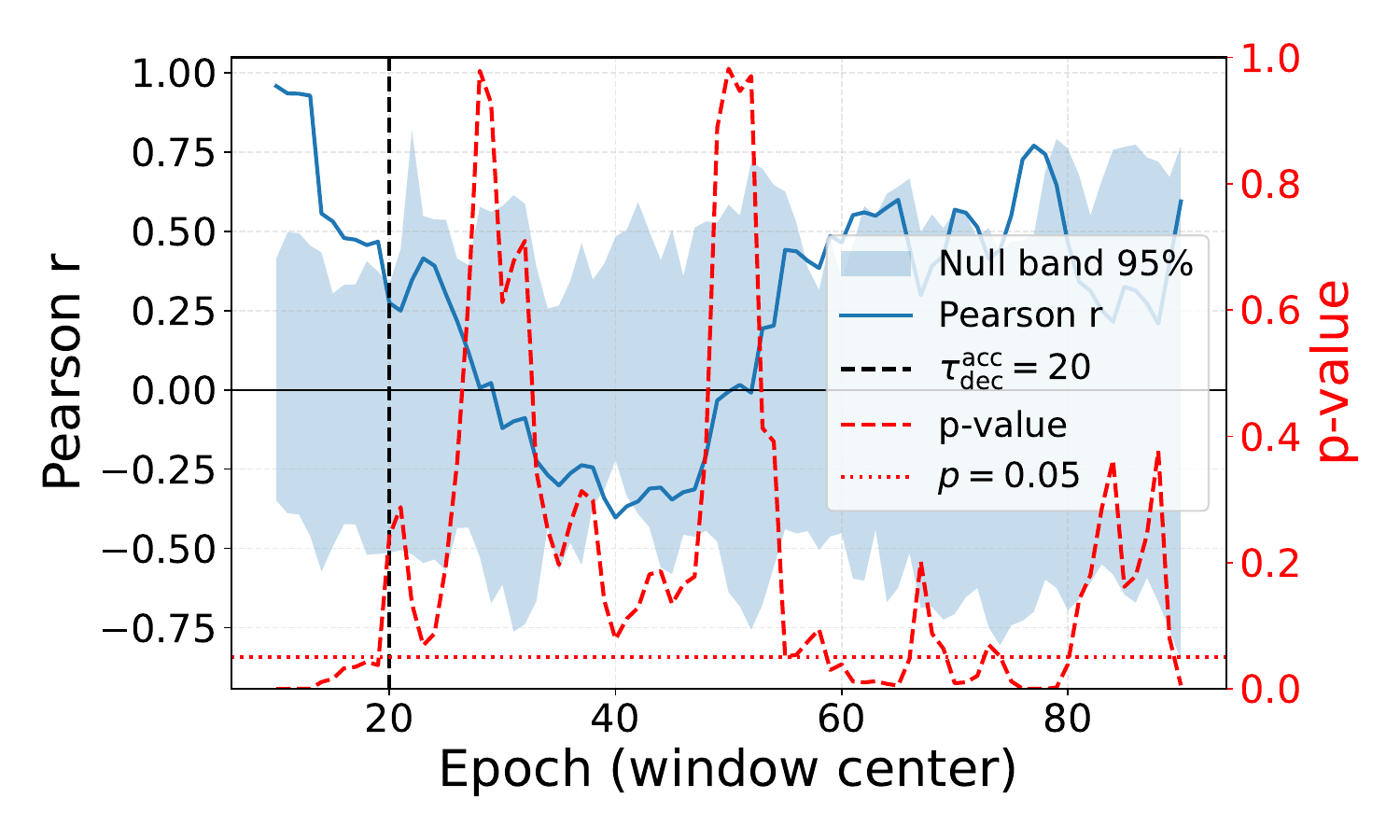}
    \caption{Sliding-window Pearson correlation $r(t)$ between the detrended accuracies of a reference trajectory and one perturbed replica, shown as a function of the window-center epoch (window length $W=20$, step $s=1$). For each window, a null distribution is generated from $n_{\mathrm{null}}=250$ nonzero circular shifts of the perturbed segment, yielding the shaded $95\%$ confidence band. The corresponding Pearson-test $p$ values are shown in red on the right axis; the horizontal dotted line marks the significance threshold $p_{\mathrm{th}}=0.05$. The vertical dashed line indicates the resulting decorrelation time $\tau_{\mathrm{dec}}^{\mathrm{acc}}$, defined as the center of the first window for which the observed correlation lies within the null confidence band, is statistically non-significant, $p(t)>p_{\mathrm{th}}$, and satisfies $|r(t)|<r_{\mathrm{th}}$. Results correspond to $\eta=10$.}
    \label{fig:example_tau_decorr_accuracy}
\end{figure}

With these disclaimers in mind, Fig.~\ref{fig:example_accuracy} illustrates the evolution of test accuracy as a function of the number of training epochs, for two initially close network initializations. The black markers also show the embedding-based separation $d(t)$ between the scalarized trajectories, while the blue and red curves show the corresponding test accuracies. At early epochs, when $d(t)\approx 10^{-6}$, both accuracy trajectories evolve almost synchronously, as expected for trajectories that remain close in the underlying parameter space dynamics. As training progresses and the distance in embedding space increases, the accuracy traces begin to desynchronize and exhibit intermittent deviations, plateaus, and crossings. Intuitively, one can then estimate from these signals an accuracy-based decorrelation time $\tau_{\mathrm{dec}}^{\mathrm{acc}}$ as the earliest epoch where the cross-correlation function between the two accuracies detects a complete loss of correlations, according to some protocol and hypothesis test.\\
%However, these fluctuations do not identify a unique or reproducible separation epoch in accuracy space. Even after $d(t)$ has clearly increased, the two accuracy curves may transiently realign or remain numerically close, reflecting the compressed and saturated nature of the observable. Therefore, although accuracy is not an ideal quantity for measuring decorrelation times, it remains one of the few experimentally observables directly linked to training performance. 
%Motivated by these considerations, we introduce an accuracy-based decorrelation time, $\tau_{\mathrm{dec}}^{\mathrm{acc}}$, together with a dedicated methodology designed to extract meaningful decorrelation information from the accuracy dynamics without relying on a simple threshold over raw accuracy differences. 
Such protocol is as follows: 
we define $\tau_{\mathrm{dec}}^{\mathrm{acc}}$ from a sliding-window correlation analysis of the reference and perturbed accuracies. Given two (accuracies) time series, $a_{\mathrm{ref}}(t)$ and $a_{\mathrm{pert}}(t)$, we consider windows of length $W$ displaced by a step $s$, and compute in each window the Pearson correlation coefficient $r(t)$ after linear detrending in order to remove slow drifts. To determine statistical significance, we construct a window-wise null model as follows: within each window, we generate $n_{\mathrm{null}}$ realizations by applying random nonzero circular shifts to the perturbed trace. This procedure preserves the temporal structure of the signal within the window while disrupting its alignment with the reference trace. From the resulting null correlations $\{r_0\}$ we extract a two-sided confidence interval $[r_{\mathrm{low}}(t),r_{\mathrm{high}}(t)]$ at a certain confidence level. A window is classified as decorrelated if (i) the observed correlation lies within this null interval, $r_{\mathrm{low}}(t)\le r(t)\le r_{\mathrm{high}}(t),$ (ii) the Pearson correlation is statistically non significant, namely $p(t)>p_{\mathrm{th}}$, and (iii) the absolute correlation is sufficiently small, $|r(t)|<r_{\mathrm{th}}$.\\ 
As an illustrative example in Fig.~\ref{fig:example_tau_decorr_accuracy}, we estimate $\tau_{\mathrm{dec}}^{\mathrm{acc}}$ by applying this method to the accuracy traces of a reference run and one of its perturbed replicas, using a window length $W=20$ epochs and a unit step $s=1$. The window length is chosen to be long enough to smooth out epoch-to-epoch fluctuations and provide a stable estimate of the correlation, while still remaining sufficiently local in time to detect the onset of decorrelation. The unit step maximizes temporal resolution, allowing $\tau_{\mathrm{dec}}^{\mathrm{acc}}$ to be identified at the earliest epoch compatible with the criterion. For each window we compute the Pearson correlation between detrended segments, generate $n_{\mathrm{null}}=250$ correlations and extract a two-sided $95\%$ confidence interval from the null distribution. We then define $\tau_{\mathrm{dec}}^{\mathrm{acc}}$ as the centered epoch of the first window for which the observed correlation is both compatible with the null interval, statistically nonsignificant with $p_{\mathrm{th}}=0.05$, and sufficiently small in absolute value, $|r(t)|<r_{\mathrm{th}}$, with $r_{\mathrm{th}}=0.50$. 
Figure~\ref{fig:example_tau_decorr_accuracy} shows a representative realization of this method, including the sliding-window correlation, the corresponding null confidence band, and the associated $p$-values. In this case, we estimate $\tau_{\mathrm{dec}}^{\mathrm{acc}}=20$ as the decorrelation time that satisfies the criteria. 

%We have presented a methodology to estimate the decorrelation time from the accuracy quantifier. However, the difficulties encountered when attempting to define a decorrelation time from accuracy suggest that standard performance observables may not be the most suitable quantities for characterizing the geometric separation in the parameter space between training trajectories. In this context, the scalar embedding becomes particularly appealing, as it provides direct access to the evolution of trajectories in a reduced space that preserves their underlying dynamical structure. 

\begin{figure}[h]
    \centering
    \includegraphics[width=1\linewidth]{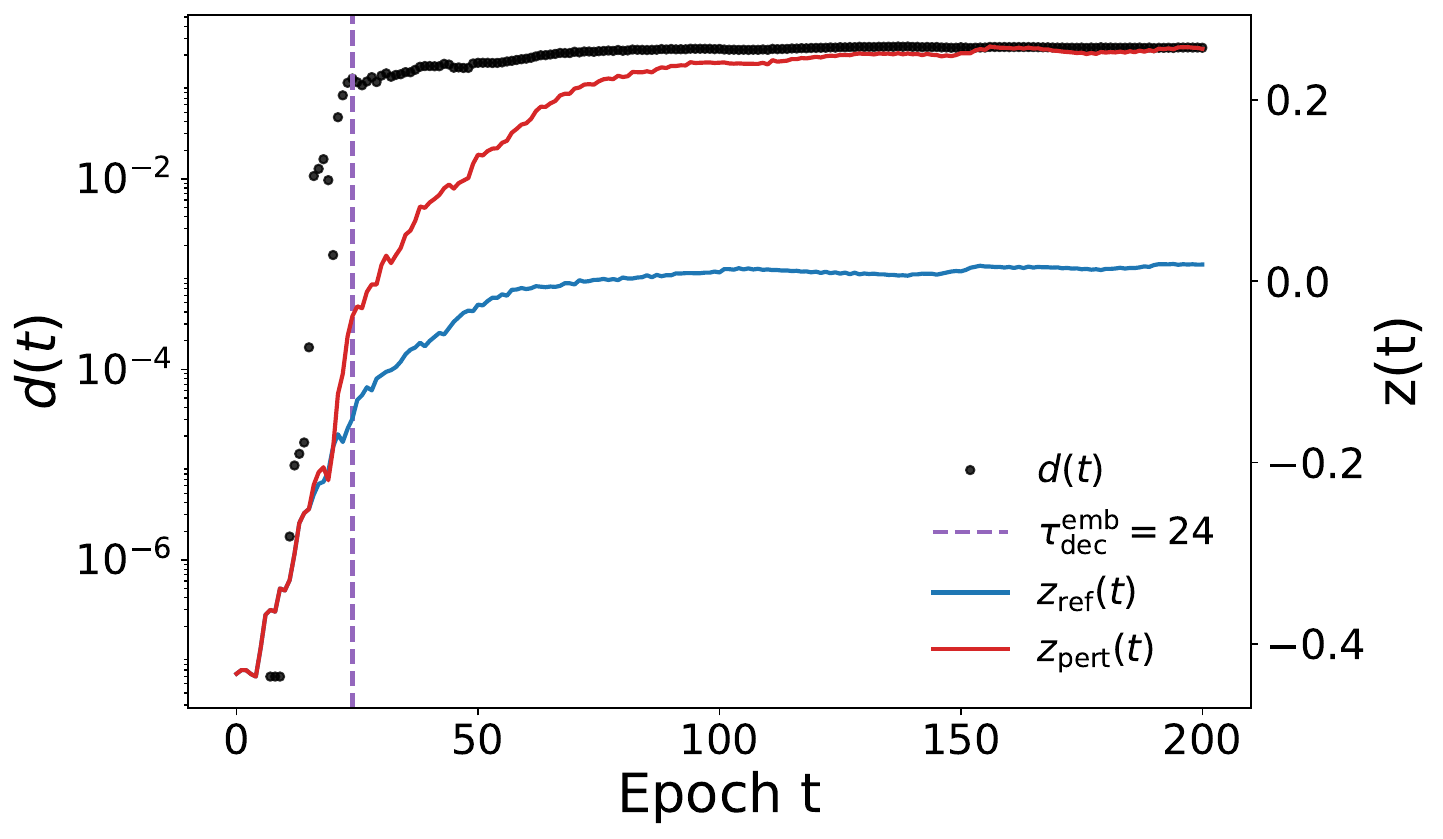}
    \caption{Example of the embedding-based decorrelation time $\tau_{\mathrm{dec}}^{\mathrm{emb}}$ for a reference trajectory and one of its perturbed replicas. The left axis shows the separation $d(t)$ between the corresponding embedded trajectories, while the vertical dashed line marks $\tau_{\mathrm{dec}}^{\mathrm{emb}}$. The right axis shows the scalar embeddings of the reference (blue) and perturbed (red) trajectories. Results correspond to $\eta=10$.}
    \label{fig:example_tau_emb}
\end{figure}

Moving on, we now consider again the scalar embedding trajectories, and attempt to define a characteristic time $\tau_{\mathrm{dec}}^{\mathrm{emb}}$ which quantifies the actual decorrelation time $\tau_{\mathrm{dec}}$. The idea here is to associate such quantity to the typical timescale needed for the exponential separation of trajectories to saturate, i.e. something akin to a Lyapunov time.
%Building on this idea, we define an embedding-based decorrelation time, $\tau_{\mathrm{dec}}^{\mathrm{emb}}$, as the first epoch at which two initially nearby trajectories become distinguishable in embedding space according to a prescribed separation criterion. 
A representative example is shown in Fig.~\ref{fig:example_tau_emb}, where initially close trajectories (blue and red) exponentially separate (black curve) until a characteristic time where such distance starts to saturate. Such change of behavior is rather easy to identify. 
%that initially overlap progressively separate as training evolves, 
%making the decorrelation time easy to identify. 
To automate the detection of $\tau_{\mathrm{dec}}^{\mathrm{emb}}$, for each initialization we consider one reference trajectory and several perturbed replicas, compute at each epoch $t$ the separation $d(t)=\left|z_{\mathrm{pert}}(t)-z_{\mathrm{ref}}(t)\right|$, and define
$\tau_{\mathrm{dec}}^{\mathrm{emb}}$ as the epoch at which the numerical time derivative of $d(t)$ in a log-linear scale drops below a threshold set to $15\%$ of its maximum value and is maintained low for a few epochs. Incidentally, we have also checked that the resulting characteristic is often a few epochs later than the one where $d(t)\geq 10^{-3}$ for the first time. 
%= \min \left\{t:\, d(t) > 10^{-3}\right\}.$ If no threshold crossing occurs within the observation window, the pair is regarded as not decorrelated on the timescale probed. This operational definition provides a direct geometric proxy for the epoch at which two initially nearby training trajectories become distinguishable in the embedding, thereby enabling systematic comparisons across perturbations, initial conditions and hyperparameter settings. 
In the example of Fig.~\ref{fig:example_tau_emb}, we find  $\tau_{\mathrm{dec}}^{\mathrm{emb}}=24$.

\begin{figure}[h]
    \centering
    \includegraphics[width=1\linewidth]{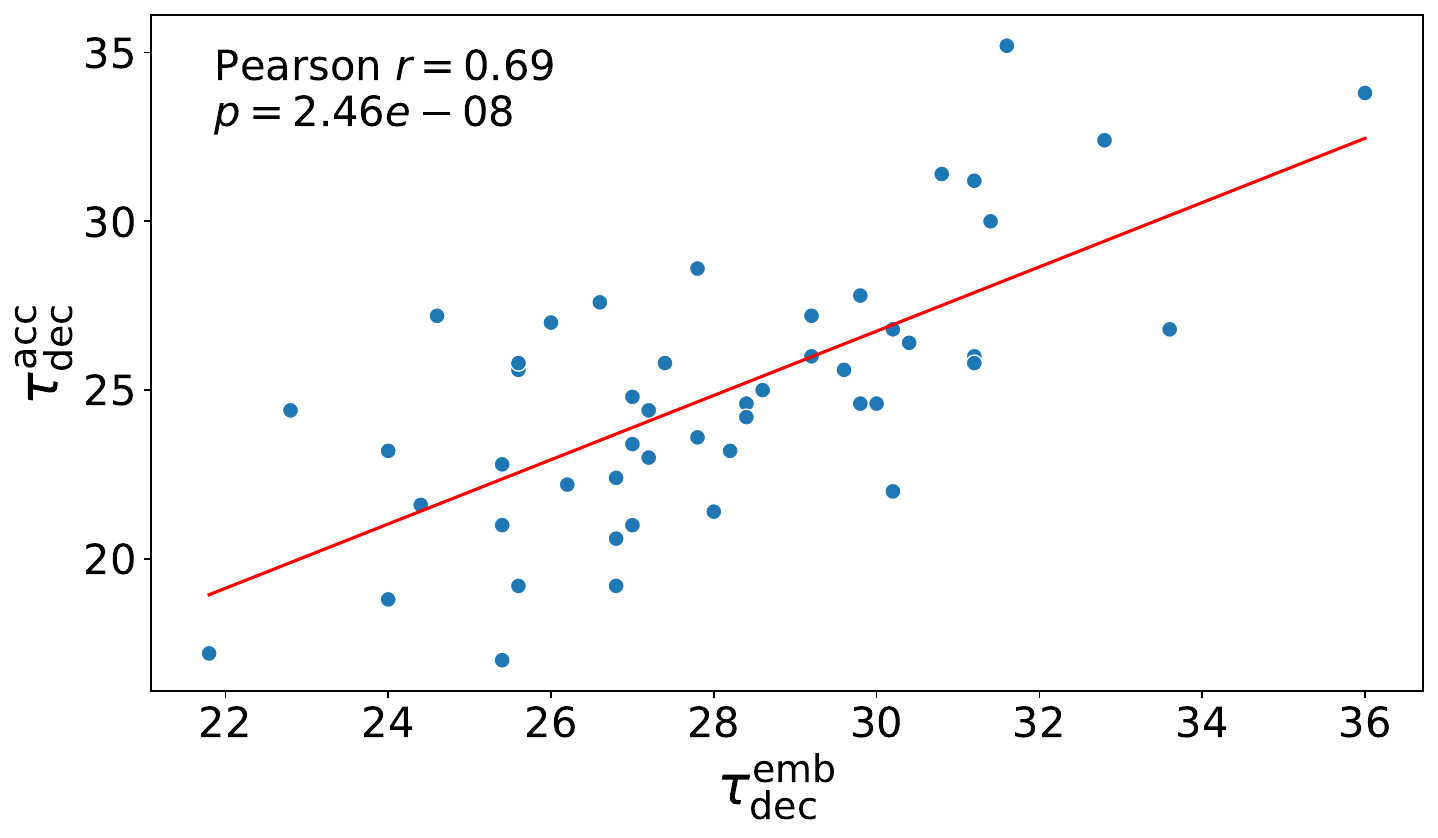}
    \caption{Scatter plot of $\tau_{\mathrm{dec}}^{\mathrm{acc}}$ versus $\tau_{\mathrm{dec}}^{\mathrm{emb}}$ across initial conditions for $\eta=10$. Each point corresponds to one initial condition, with both decorrelation times averaged over the five perturbations for each initial condition.}
    \label{fig:correlation_tau_decorr}
\end{figure}

We have applied such procedure to all pairs of initially close trajectories, for all initial conditions. In order to compare the estimation of decorrelation time via the embedding with respect to the one found via the accuracy trajectories, 
Fig.~\ref{fig:correlation_tau_decorr} scatter plots $\tau_{\mathrm{dec}}^{\mathrm{acc}}$ vs $\tau_{\mathrm{dec}}^{\mathrm{emb}}$. The two quantities are clearly linearly correlated in a statistically significant way (Pearson $r=0.69$, p-value$=2.5\times 10^{-8}$).
%Having defined two methodologies to estimate decorrelation times, $\tau_{\mathrm{dec}}^{\mathrm{emb}}$ and $\tau_{\mathrm{dec}}^{\mathrm{acc}}$, it is natural to ask to what extent they reflect a common underlying dynamical timescale. To address this question, we compute the two estimators across different initial conditions for $\eta=10$. A positive correlation would indicate that the epoch at which accuracies lose mutual coherence still retains information about the onset of geometric separation in parameter space. In Fig.~\ref{fig:correlation_tau_decorr}, we present the values of both decorrelation times for each initial condition. We find a moderate Pearson correlation between the two estimators, $r\simeq 0.7$, together with a p-value of order $10^{-8}$. 
%This suggests that both estimators capture the same variability in sensitivity to perturbations across different initial conditions, despite being constructed from very different dynamical representations.

Furthermore, we argue that the correlation coefficient is not higher precisely because estimation of decorrelation times via accuracy trajectories has some problems --as identified earlier--. 
%this is expected, since the compressed and noisy nature of the accuracy observable makes the identification of decorrelation substantially less precise than in the scalar embedding. In the original parameter space, the divergence between trajectories evolves in a space of tens of thousands of dimensions, making its geometric structure essentially inaccessible to direct visualization. 
%By contrast, the scalar embedding maps the dynamics onto a reduced representation in which the separation between nearby trajectories becomes immediately observable and straightforward to quantify. As a consequence, $\tau_{\mathrm{dec}}^{\mathrm{emb}}$ is less affected by the coarse-grained nature of the accuracy, as well as by methodological choices such as the window length required to define $\tau_{\mathrm{dec}}^{\mathrm{acc}}$. In particular, the embedding can reveal persistent divergence even in regimes where the corresponding accuracies remain nearly indistinguishable, for instance along broad plateaus or close to performance saturation, where estimates of $\tau_{\mathrm{dec}}^{\mathrm{acc}}$ become highly variable. 
Overall, the embedding-based definition yields a more stable and easier-to-estimate proxy for the decorrelation time.
%, allowing a clearer assessment of sensitivity to initial perturbations and more reliable comparisons across initial conditions and hyperparameter settings.

\subsection{Distribution of stationary points in the parameter landscape}
\begin{figure}[h]
    \centering
    \includegraphics[width=1\linewidth]{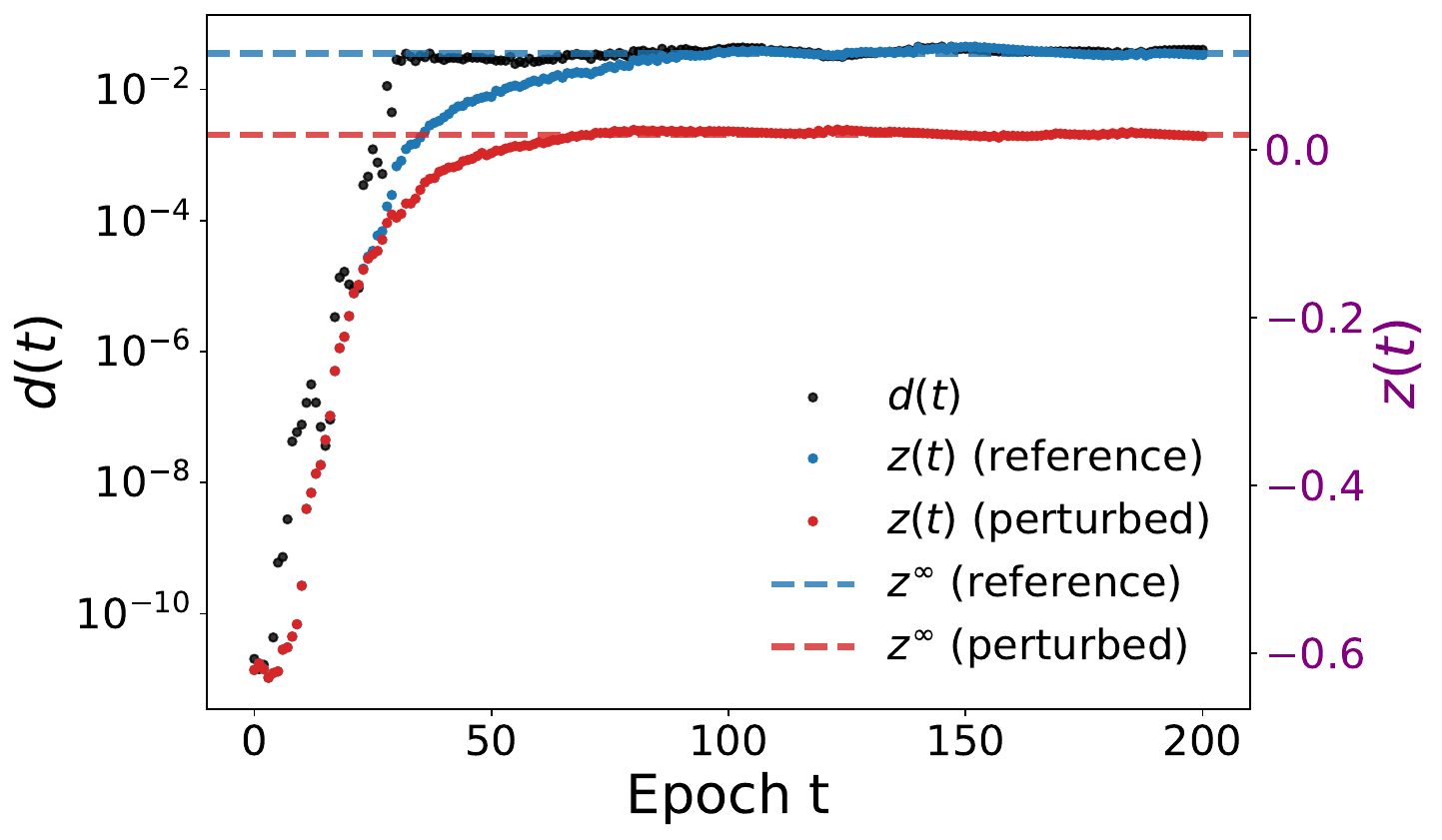}
    \caption{Example of asymptotic stabilization in the scalar embedding. The black marker represents the distance $d(t)$ between the embedded reference and perturbed trajectories as a function of training epoch $t$ (left axis). The blue and red curves show the corresponding scalar embeddings $z(t)$ (right axis), and the dashed horizontal lines mark their asymptotic values ($z^{\infty}$). Results correspond to $\eta=10.0$. }
    \label{fig:embedding_stabilization}
\end{figure}

%We now investigate how minima are organized within the scalar embedding. We assume that the asymptotic states reached by the trajectories along the embedding axis provide a coarse-grained representation of the corresponding minima.
In the context of ANN training, the loss is highly non-convex and its landscape is populated by very many local minima, whose distribution is often difficult to characterize due to the curse of dimensionality. 
Here we aim to understand how such distribution of minima is mapped into the scalar embedding. 
%highly non-convex learning dynamics can be interpreted as an optimization process over a high-dimensional loss landscape, where the final trained configurations correspond to minima of the loss function. We assume that the asymptotic states reached by the trajectories along the embedding axis provide a coarse-grained representation of the corresponding minima. 
To that aim, for each scalar trajectory $z(t)$ we define its asymptotic, quasi-stationary state as $z^{\infty} := \left\langle z(t) \right\rangle_{t \in \mathcal{T}_{\mathrm{final}}}$ where $\mathcal{T}_{\mathrm{final}}$ denotes a window consisting of the last 50 training steps. Empirical observation indeed shows that, in the scalar embedding, after a transient trajectories often stabilise within a stationary state, see Fig.~\ref{fig:embedding_stabilization} for an illustration.
%We say that the scalar trajectory has reached a stationary state
%when the separation between a reference trajectory and a perturbed one becomes stationary. As illustrated in Fig.~\ref{fig:embedding_stabilization}, the distance $d(t)$ initially grows exponentially, reflecting the divergence of nearby trajectories for $\eta=10$. Following this phase, $d(t)$ saturates and fluctuates around a constant value, indicating that the relative separation between the two embedded trajectories has stabilized. 
%Once this saturation occurs, both scalar embeddings evolve in parallel, converging to asymptotic values that remain approximately constant throughout subsequent training epochs as shown in Fig.~\ref{fig:embedding_stabilization}. This behavior suggests that the optimization dynamics have relaxed into a stable region of the loss landscape, such that the embedded trajectories have reached their asymptotic positions. 
We identify the onset of a plateau in the separation distance between initially close trajectories $d(t)$ as the starting point of the stabilization of the scalar trajectories, and use the subsequent epochs to estimate the asymptotic value $z^{\infty}$. 

\begin{figure}[h]
    \centering
    \includegraphics[width=1\linewidth]{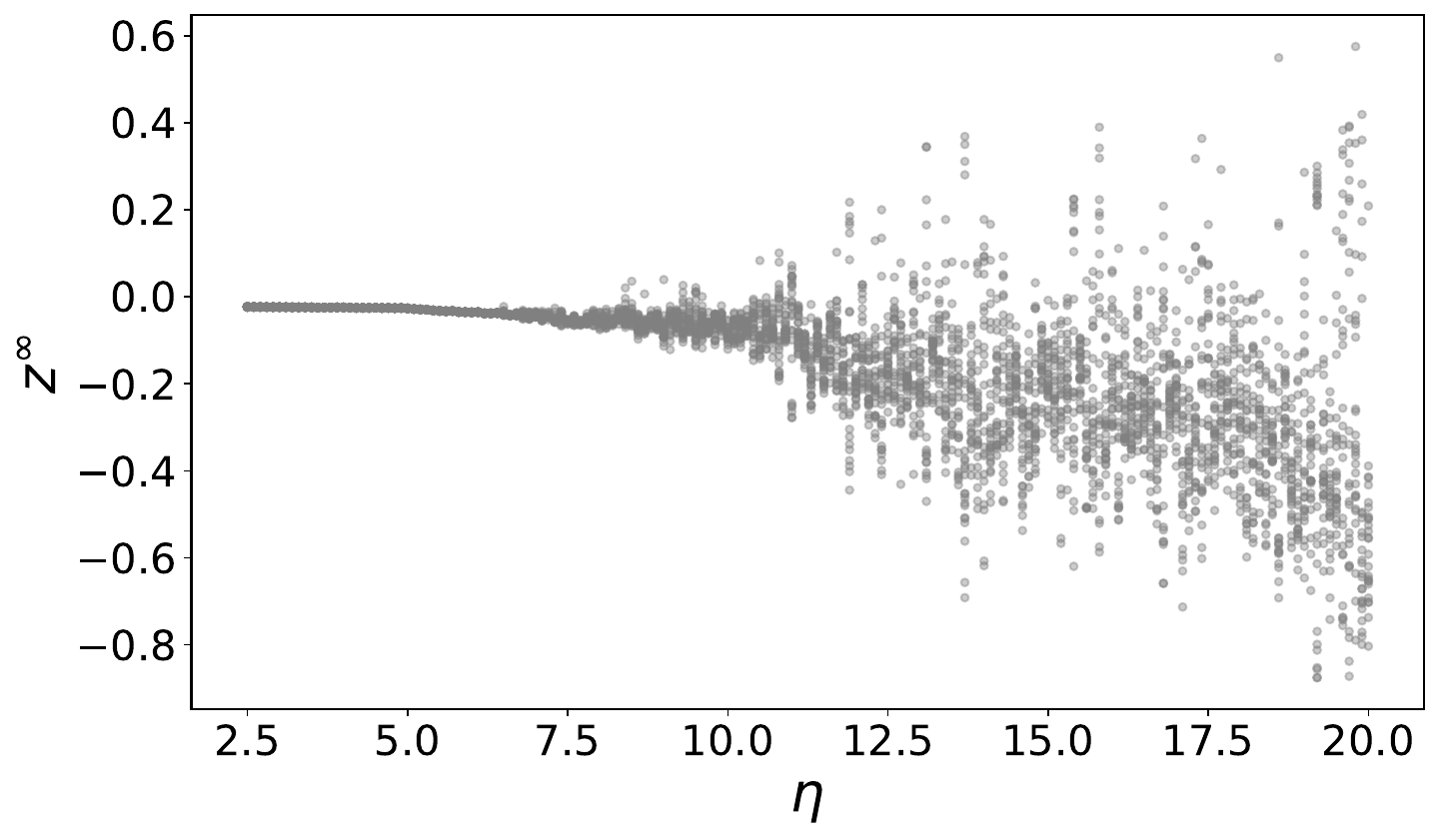}
    \caption{Asymptotic embedded states as a function of the learning rate. Each point represents an asymptotic embedding value $z^{\infty}$ obtained from an individual training trajectory, with 26 trajectories analyzed for each value of $\eta$ (one reference trajectory and 25 perturbed realizations). For small learning rates, the asymptotic states are confined to a narrow region of the embedded space, indicating a relatively compact organization of final states. As $\eta$ increases, the spread of $z^{\infty}$ broadens significantly, revealing a progressively more heterogeneous organization of the asymptotic states in the embedded landscape.} %\textcolor{red}{COMENTAR QUE NO ES UN SIMPLE PERIOD DOUBLING ROUTE TO CHAOS QUE ES ALGO MAS COMPLICADO DE HECHO HAY Q ACLARAR Q AQUI ESTAMOS PINTANDO PUNTOS ESTACIONARIOS (PUNTOS FIJOS?), NO SON CICLOS COMO EN LA REPRESENTACION TIPICA...}}
    \label{fig:barrido_eta}
\end{figure}

\medskip \noindent Note that, in order to be able to capture true local minima and to be able to sample a representative amount of them, we need to track trajectories that (i) by construction can initially spread over the whole loss landscape, and (ii) after such initial exploration phase, can eventually relax into local minima. Such interplay requires to fix a learning rate $\eta$ of the optimizer that indeed guarantees trajectories balancing such exploration/exploitation tradeoff \cite{jimenez2026leveraging}. In Fig.~\ref{fig:barrido_eta} we plot, for different values of $\eta$, the set of values $\{z^{\infty}\}$ found from the scalar embedding of the original (training) network trajectories. In particular,
%Having defined the asymptotic embedded value $z^{\infty}$ for each trajectory, we now examine how the set of asymptotic states accessed during training depends on the learning rate. 
%To this end, 
for each value of $\eta$ we consider one reference trajectory together with 25 perturbed replicas generated from the same initial condition, and compute the corresponding asymptotic values $z^{\infty}$. This yields, for each learning rate, a set of 26 asymptotic endpoints, $\{z_0^{\infty}, z_1^{\infty}, \ldots, z_{25}^{\infty}\}$ which are shown as a function of $\eta$ in Fig.~\ref{fig:barrido_eta}.\\
For $\eta \lesssim 7.5$, these asymptotic values remain confined to a relatively narrow interval, indicating that nearby trajectories cannot really explore much of the loss landscape. As $\eta$ increases, however, the spread of $z^{\infty}$ broadens substantially, revealing a growing diversity of asymptotic outcomes and a stronger sensitivity on small perturbations of the initial condition. In particular, a marked increase in the variability of the asymptotic endpoints appears around $\eta \approx 7.5$. This range coincides with the regime in which the network attains its best training-time performance in our previous experiments as shown in Fig~\ref{fig:tau}. The results in Fig.~\ref{fig:barrido_eta} are consistent with the interpretation that, in this sweet spot, the optimization dynamics explore a broader set of regions in parameter space, ultimately giving rise to a more heterogeneous distribution of asymptotic embedded states.

\begin{figure}[h]
    \centering
    \includegraphics[width=1\linewidth]{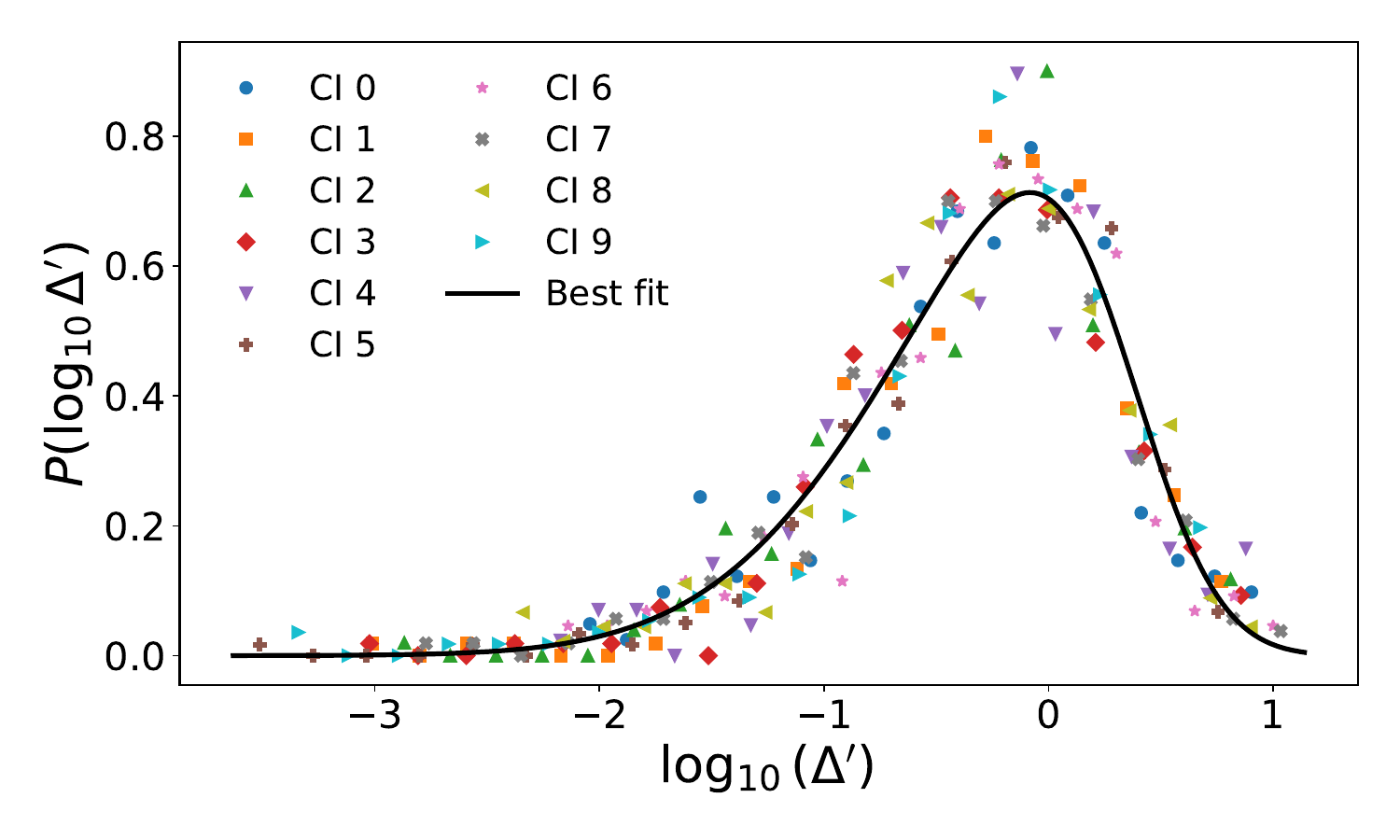}
    \caption{Rescaled spacing distributions for $\eta={10}$. Markers corresponding to 10 independent initial conditions (CI 0–9), shown in terms of the transformed variable $\log_{10}(\Delta')$, with $\Delta'=\Delta/\langle\Delta\rangle$. Each set of markers represents the empirical histogram of a single initial condition. The solid line shows the skew-normal fit to the pooled data, yielding fitted parameters $\alpha=-2.78$, $\xi=0.36$, and $\omega=0.91$.}
    \label{fig:reescalado}
\end{figure}

\medskip \noindent
Once we have located the value of $\eta=10$ that a priori guarantees that the stationary points $\{z^{\infty}\}$
are (i) local minima, and that these are (ii) scattered throughout a considerable region of the loss landscape, we are ready to characterise its distribution along the one-dimensional axis of the scalar embedding. Instead of characterising its raw distribution, we focus on the distribution of nearest-neighbor spacing $\Delta$.
%These observations motivate the introduction of the spacing observable $\Delta$, which provides a quantitative measure of how densely the asymptotic endpoints are arranged along the embedding axis. 
In practice, starting from one reference trajectory and $M$ perturbed replicas, denoted by $\{z_{0}(t), z_{1}(t), \ldots, z_{M}(t)\}$, we obtain the corresponding set of asymptotic values  $\{z_{0}^{\infty}, z_{1}^{\infty}, \ldots, z_{M}^{\infty}\}$. Without loss of generality, we can then order these values as
\begin{equation}
    z_{(0)}^{\infty} > z_{(1)}^{\infty} > \cdots > z_{(M)}^{\infty}
\end{equation}
We then define the set of nearest-neighbor spacings between consecutive asymptotic states as 
\begin{equation}
    \Delta = \{\Delta_1, \Delta_2, \ldots, \Delta_M\},
\qquad
\Delta_i = z_{(i-1)}^{\infty} - z_{(i)}^{\infty}.
\end{equation}
The spacing sequence $\Delta$ provides a quantitative description of how the asymptotic states are arranged along the scalar embedding axis. In particular, it captures the local density of stationary endpoints: small values of $\Delta_{i}$ indicate closely packed asymptotic states, whereas large values signal more isolated ones. In this way, $\Delta$ offers a simple observable with which to characterize the fine organization of the minima explored by training in the embedded landscape.\\ 
%Having introduced the spacing observable $\Delta$, we next study its statistical properties at fixed learning rate $\eta$. 
For $\eta=10$, we generate an ensemble of training trajectories from $N$ independent initial conditions, each supplemented with $M$ small perturbations. For every realization, we compute the associated asymptotic embedded values $z^{\infty}$, sort them, and extract the spacings $\Delta_{i}$ between consecutive endpoints. Pooling these spacings over all realizations yields the empirical distribution $P(\Delta)$ corresponding to that learning rate. This distribution allows us to move beyond individual trajectories and probe whether the arrangement of nearby minima in the embedded landscape exhibits statistical regularities.\\ 
A first question is whether these spacing statistics vary for different initial conditions. As a matter of fact, remind that different regions of the parameter landscape can have different finite Lyapunov exponents, and therefore the initial separation of nearby trajectories can be more rapidly amplified in some regions than in others, eventually resulting in spacing distributions whose scale could be different.
We indeed find that the raw distributions $P(\Delta)$ are not fully compatible across initial conditions, and that there is a clear scale effect. However, after rescaling the spacings in each realization by their mean value, $\Delta'=\frac{\Delta}{\langle \Delta \rangle}$, %and applying the logarithmic transformation $\log_{10} \Delta' $, 
the resulting distribution becomes independent of the region of the parameter space from which the trajectories start. In particular, we sampled 10 initial conditions, and built 250 perturbations around each of the 10 initial conditions. These result in 10 different spacing distributions. When adequately rescaled, these distributions were all compatible ($10$-sample Anderson-Darling test \cite{scholz1987k}, p-value $ \simeq 0.93$). The resulting collapse of the rescaled distributions is shown in Fig~\ref{fig:reescalado} (observe that we are plotting $\log_{10} \Delta'$).\\

\medskip \noindent Once we made sure that the statistic $\Delta'$ is not conditioned on the initial condition, we can then merge all 10 sets of 250 points to build a joint sample of 2500 values. We then tested the empirical pdf of $\log_{10} \Delta'$ against a
%We then examine that common shape in more detail. Since the distributions of $\Delta'$ are statistically compatible across initial conditions, we pool the rescaled spacings from the ten realizations, obtaining a joint sample of 2500 values. 
%We analyze the logarithmically transformed variable $\log_{10} \Delta'$ and fit a 
set of candidate parametric distributions, including the normal, skew-normal, Student-t, Laplace, logistic, and generalized normal distributions, using Maximum Likelihood Estimation. In particular, the probability distribution function of a skew-normal distribution is given by
%\begin{equation}
%    f(x;\xi,\omega,\alpha) = \frac{2}{\omega}\phi \left(\frac{x-\xi}{\omega}\right) \Phi \left(\alpha\left(\frac{x-\xi}{\omega}\right)\right)
%\end{equation}
%\begin{widetext}
%\begin{equation}
%    f(x;\xi,\omega,\alpha) = 
%    \frac{1}{\omega \sqrt{2\pi}}
%    \exp \left[-\frac{1}{2}\left(\frac{x-\xi}{\omega}\right)^{2}\right]
%    \left[1 + \operatorname{erf}\left(\alpha\left(\frac{x-\xi}{\omega\sqrt{2}}\right)\right) \right]
%\end{equation}
%\end{widetext}
%\begin{equation}
%    f(x;\xi,\omega,\alpha) = 
%    \frac{1}{\omega \sqrt{2\pi}}
%    e^{-z^{2}/2 }
%    \left[1 + \operatorname{erf}\left(\frac{\alpha z}{\sqrt{2}}\right) \right]
%\end{equation}
%\begin{equation}
%    f(x;\xi,\omega,\alpha) = 
%    \frac{1}{\omega \sqrt{2\pi}}
%    \exp \left[-\frac{1}{2}\left(\frac{x-\xi}{\omega}\right)^{2}\right]
%    \left[1 + \operatorname{erf}\left(\alpha\left(\frac{x-\xi}{\omega\sqrt{2}}\right)\right) \right]
%\end{equation}
%\begin{equation}
%\scalebox{0.85}{$f(x;\xi,\omega,\alpha) = 
%    \frac{1}{\omega \sqrt{2\pi}}
%    \exp \left[-\frac{1}{2}\left(\frac{x-\xi}{\omega}\right)^{2}\right]    \left[1 + \operatorname{erf}\left(\alpha\left(\frac{x-\xi}{\omega\sqrt{2}}\right)\right) \right]$}
%\end{equation}
%\begin{equation}
%    f(x;\xi,\omega,\alpha) = 
%    \frac{e^{-\frac{1}{2}\left(\frac{x-\xi}{\omega}\right)^{2}}}{\omega \sqrt{2\pi}}
%    \left[1 + \operatorname{erf}\left(\alpha\left(\frac{x-\xi}{\omega\sqrt{2}}\right)\right) \right]
%\end{equation}
\begin{equation}
    f(x;\xi,\omega,\alpha) = 
    \frac{e^{-\frac{1}{2}\left(\frac{x-\xi}{\omega}\right)^{2}}}{\omega \sqrt{2\pi}}
    \left[1 + \operatorname{erf}\left(\frac{\alpha \,(x-\xi)}{\omega\sqrt{2}}\right) \right]
\end{equation}
where $x = \log_{10}(\Delta')$, $\xi$ is a location parameter, $\omega>0$ a scale parameter, $\alpha$ controls the skewness and $\operatorname{erf}$ denotes the error function. %, and $\phi$ and $\Phi$ denote the standard normal pdf and cdf.
Goodness of fit was evaluated using a one-sample Kolmogorov-Smirnov test. Among the candidate distributions considered, only the skew-normal distribution was not rejected at the 5\% significance level. The resulting fit is shown in Fig.~\ref{fig:reescalado}, where the ten rescaled distributions are displayed individually together with the fitted probability density function. %nuevo 
We find that this statistical structure is characteristic of the transition-to-chaos region, where training trajectories balance exploration of the landscape with eventual convergence to local minima. As expected, for small learning rates the dynamics probe only a limited portion of the landscape, while for very large learning rates trajectories explore broadly but fail to converge to well-defined minima, preventing a meaningful characterization of the corresponding minima distribution. 

\medskip \noindent 
The interpretation of these results is that the variability observed across the raw spacing distributions $\Delta$ is primarily controlled by differences in the local instability properties associated with each initial condition, due to different local Lyapunov exponents that lead to nonuniform expansion properties of the dynamics along different regions of phase space. 
%This behavior already appears in simple systems such as the fully chaotic logistic map, $x_{t+1} = 4x_{t}(1-x_{t})$, where local estimates of the expansion rate exhibit a nontrivial distribution around the asymptotic value $\lambda=\ln2$. 
%A similar mechanism may be operating here. Initial conditions associated with larger local Lyapunov exponents are expected to amplify perturbations more rapidly during the early stages of training, potentially leading to larger characteristic separations between the corresponding asymptotic embedded states. Conversely, trajectories with smaller local instability would naturally generate smaller average spacings. From this perspective, 
We speculate that the mean spacing $\langle \Delta \rangle$ associated to a given initial condition may indeed be related to the local Lyapunov exponents in that region.
%encode the characteristic expansion scale associated with the local dynamical instability of that initial condition. 
Under this interpretation, the collapse of the rescaled distributions after normalizing by $\langle \Delta \rangle$ suggests that most of the variability across initial conditions can be attributed to this scale factor, which is indeed independent of the underlying loss geometry. The remaining distributional shape, which becomes approximately universal after rescaling, would then fully reflect said geometry. The precise shape of the rescaled distribution appears to be a skewed lognormal distribution. How this seemingly universal shape is connected to the specific loss landscape geometry in the original, high-dimensional space and how this might depend on different neural network architectures and regression/classification tasks is a fascinating open problem.

%In other words, the organization of minima in the embedded landscape may be controlled by two distinct ingredients: a scale-setting mechanism related to the instability of the dynamics, and a geometric \PJ{igual puede ser por la arquitectura de la red, o hiperparámetros, o es estocástico? no se bien como definir esto si geometric es adecuado} mechanism responsible for the dispersion of the spacings around that scale. 

\section{Conclusions}\label{sec:conclusion}
In this work we find that our scalar embedding technique provides a low-dimensional representation of neural network training dynamics while preserving nontrivial dynamical information from the original high-dimensional parameter space. Our representative examples show that despite compressing trajectories evolving in a space of order $5\cdot 10^{4}$ dimensions into a single scalar observable, the embedding successfully reproduces the qualitative dependence of the dynamics on the learning rate, including the emergence of sensitivity to initial conditions in the same training regimes identified in the full parameter space analysis.\\
The embedding representation also enables a direct geometric characterization of trajectory separation during training. In particular, we introduce an embedding-based decorrelation time, $\tau_{\mathrm{dec}}^{\mathrm{emb}}$, which provides a robust and interpretable measure of the timescale over which nearby trajectories lose coherence. Compared with an accuracy-based estimator, the embedding-based approach avoids the limitations associated with compressed and saturated performance observables, allowing trajectory divergence to be identified more clearly and systematically.\\
Beyond characterizing trajectory divergence, the scalar embedding also reveals statistical structure in the organization of asymptotic states in the embedded landscape. By studying the spacing observable $\Delta$ between asymptotic embedded states, we showed that the corresponding spacing distributions become statistically compatible across initial conditions after rescaling by their mean value. This collapse suggest the existence of an approximately universal underlying organization of minima in the embedded landscape, with the characteristic scale of the spacings controlled by local instability properties of the dynamics.\\
More broadly, these results support the interpretation of ANN training as a high-dimensional dynamical process whose geometric and chaotic properties can nevertheless be captured through low-dimensional representations \cite{lacasa2025scalar}. The methodology introduced here opens the possibility of studying optimization dynamics, sensitivity to perturbations, and landscape organization in neural networks using tools and concepts from dynamical systems and statistical physics in a simplified setting.\\
Possible future directions include extending the analysis to stochastic optimization methods such as stochastic gradient descent (SGD) and to alternative neural networks architectures. Another interesting direction would be to investigate whether the observed spacing statistics exhibit universal behavior across different architectures, datasets and optimization methods.\\

\noindent {\bf Acknowledgments --} PJ acknowledges funding from Maria de Maeztu (MdM) Seal of Excellence  (CEX2021-001164-M) via the FPI programme (grant PRE2022-104148), funded by the MICIU/AEI/10.13039/501100011033.
LL acknowledges partial support from project CSxAI (PID2024-157526NB-I00) funded by MICIU/AEI/10.13039/501100011033/FEDER, UE. MCS and LL acknowledge partial support from a Maria de Maeztu grant (CEX2021-001164-M) funded by the MICIU/AEI/10.13039/501100011033, and from the European Commission Chips Joint Undertaking project No. 101194363 (NEHIL).\\

\noindent {\bf Code availability --} The code used to run the simulations will be available after publication at \url{https://github.com/pedrojg8}.\\

\appendix
\section{Training efficiency and sensitivity to initial conditions}\label{sec:appendix}

\begin{figure}[h]
    \centering
    \includegraphics[width=1\linewidth]{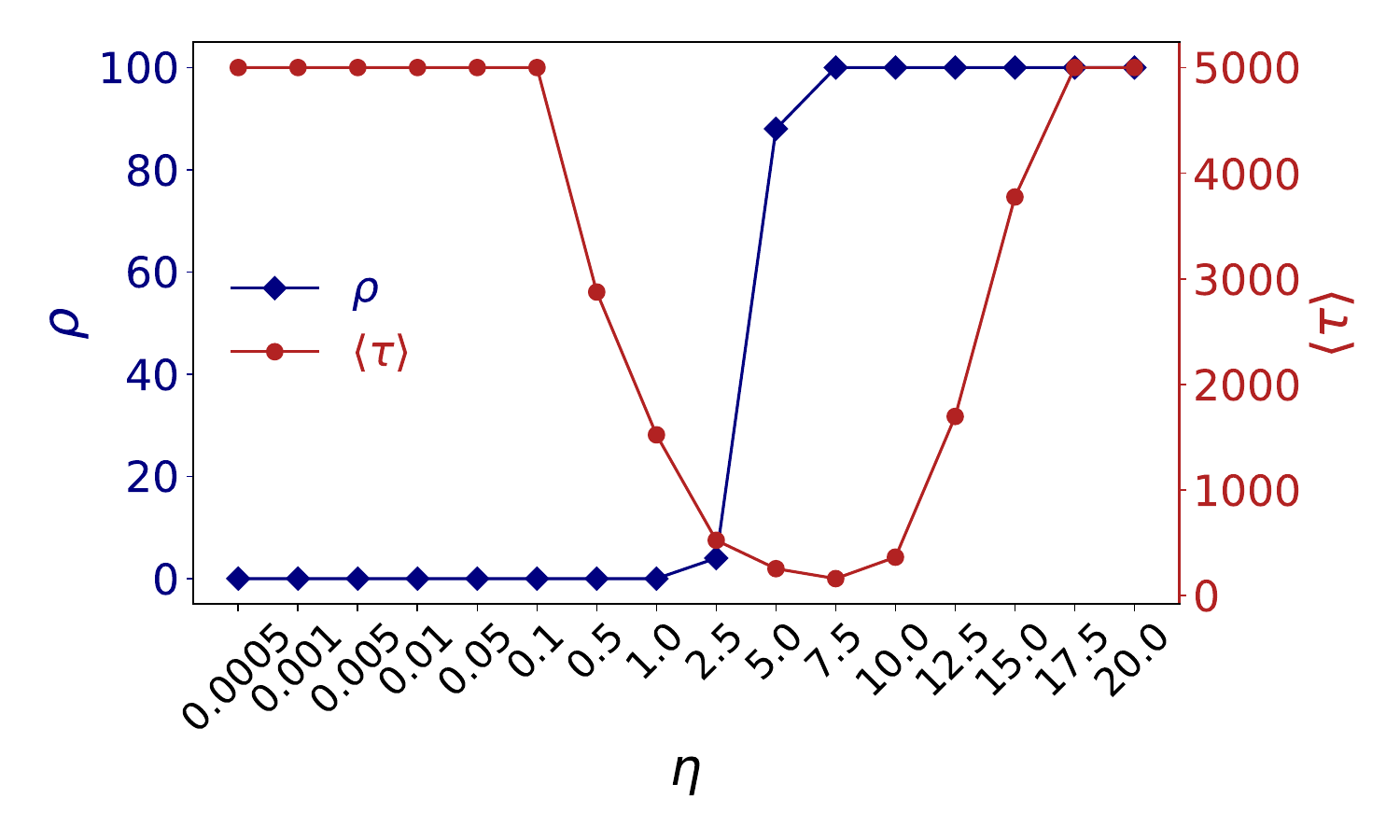}
    \caption{Percentage of positive local Lyapunov exponents, $\rho$ (left axis), and average number of epochs $\langle \tau \rangle$ required to reach $90\%$ validation accuracy (right axis) as functions of the learning rate $\eta$. The minimum of $\langle \tau \rangle$ occurs in the same learning rate region where $\rho$ approaches $100\%$}
    \label{fig:tau}
\end{figure}

To assess the relation between sensitivity to perturbations and learning performance, in \cite{jimenez2026leveraging} we computed the average number of epochs, $\langle \tau \rangle$, required for the network to reach a validation accuracy of at least $90\%$, averaged over all initial conditions. Figure~\ref{fig:tau} shows $\langle \tau \rangle$ and $\rho$, the percentage of initial conditions for which the local MLE is positive, as functions of the learning rate $\eta$.

The dependence of $\langle \tau \rangle$ on $\eta$ is markedly non-monotonic. While small learning rates lead to slow convergence, excessively large learning rates result in unstable training. Between these two extremes, $\langle \tau \rangle$ reaches a pronounced minimum around $\eta \approx 7.5$. Remarkably, this minimum coincides with the regime where $\rho$ approaches $100\%$, indicating that sensitivity to initial conditions becomes widespread across the ensemble. This observation reveals a learning rate region in which strong dynamical instability coexists with near-optimal training efficiency, motivating the focus on this regime throughout the present work.

\bibliographystyle{unsrt}
\bibliography{sample}

\end{document}